\def\year{2022}\relax
\documentclass[letterpaper]{article} 
\usepackage{aaai22}  
\usepackage{times}  
\usepackage{helvet}  
\usepackage{courier}  
\usepackage[hyphens]{url}  
\usepackage{graphicx} 
\usepackage{natbib}  
\usepackage{caption} 
\DeclareCaptionStyle{ruled}{labelfont=normalfont,labelsep=colon,strut=off} 
\usepackage{algorithm}
\usepackage{algorithmic}
\usepackage[linewidth=1pt]{mdframed}

\usepackage{newfloat}
\usepackage{listings}
\floatstyle{ruled}
\newfloat{listing}{tb}{lst}{}
\floatname{listing}{Listing}

\usepackage[utf8]{inputenc} 
\usepackage[T1]{fontenc}    
\usepackage{url}            
\usepackage{booktabs}       
\usepackage{amsfonts}       
\usepackage{nicefrac}       
\usepackage{microtype}      
\usepackage{xcolor}         
\usepackage{amsmath}
\usepackage{amssymb}
\usepackage{graphicx}
\usepackage{subcaption}
\usepackage{xspace}
\usepackage{multirow}
\usepackage{wrapfig}
\usepackage{comment}
\usepackage{amsthm }
\usepackage{mathtools, cuted}

\usepackage{wrapfig,booktabs}

\newcommand{\eg}{\textit{e.g.~}}
\newcommand{\ie}{\textit{i.e.~}}
\newcommand{\tinyimagenet}{TinyImageNet\xspace}
\newcommand{\infoNCE}{InfoNCE\xspace}
\newcommand{\RINCEsingle}{RINCE-uni\xspace}
\newcommand{\RINCEin}{RINCE-in\xspace}
\newcommand{\RINCEout}{RINCE-out\xspace}
\newcommand{\RINCEoutin}{RINCE-out-in\xspace}
\newcommand{\cifarhun}{Cifar-100\xspace}
\newcommand{\cifarten}{Cifar-10\xspace}
\newcommand{\supmat}{Sup.~Mat.\xspace}

\DeclareMathOperator{\cossim}{cos\_sim}

\title{Ranking Info Noise Contrastive Estimation: Boosting Contrastive Learning via Ranked Positives}

\author{
    David T. Hoffmann\equalcontrib\textsuperscript{\rm ,\rm 1,\rm 2}\,
    Nadine Behrmann\equalcontrib\textsuperscript{\rm ,\rm 1}\,
    Juergen Gall\textsuperscript{\rm 3}\,
    Thomas Brox\textsuperscript{\rm 2}\,
    Mehdi Noroozi\textsuperscript{\rm 1}\,
}
\affiliations{
    \textsuperscript{\rm 1}Bosch Center for Artificial Intelligence \qquad
    \textsuperscript{\rm 2}University of Freiburg \qquad
    \textsuperscript{\rm 3}University of Bonn
}

\begin{document}

\maketitle

\begin{abstract}
This paper introduces Ranking Info Noise Contrastive Estimation (RINCE), a new member in the family of \infoNCE losses that preserves a ranked ordering of positive samples. In contrast to the standard \infoNCE loss, which requires a strict binary separation of the training pairs into similar and dissimilar samples, RINCE can exploit information about a similarity ranking for learning a corresponding embedding space. We show that the proposed loss function learns favorable embeddings compared to the standard \infoNCE whenever at least noisy ranking information can be obtained or when the definition of positives and negatives is blurry. We demonstrate this for a supervised classification task with additional superclass labels and noisy similarity scores. Furthermore, we show that RINCE can also be applied to unsupervised training with experiments on unsupervised representation learning from videos. In particular, the embedding yields higher classification accuracy, retrieval rates and performs better in out-of-distribution detection than the standard \infoNCE loss.
\end{abstract}

\begin{figure*}[t]
\centering
\includegraphics[width=0.8\linewidth]{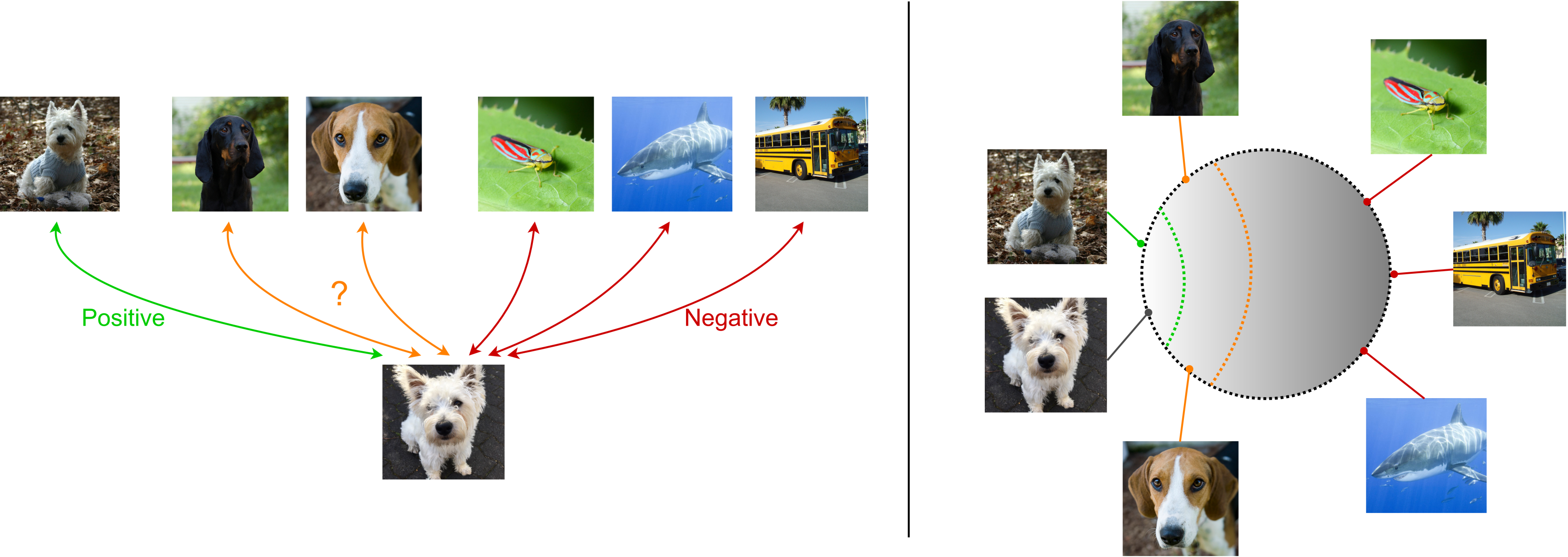}
\caption{
{\bf Contrastive Learning should not be binary.} In many scenarios a strict separation of samples in ``positives'' and ``negatives'' is not possible. So far, this grey zone (left) was neglected, leading to sub-optimal results. We propose a solution to this problem, which embeds same samples very close and similar samples close in the embedding space (right).
}
\label{fig:ranking}
\end{figure*}

\section{Introduction}
Contrastive learning recently triggered progress in self-supervised representation learning. 
Most existing variants require a strict definition of positive and negative pairs used in the \infoNCE loss or simply ignore samples that can not be clearly classified as either one or the other \cite{Zhao:2021:WMI}. Contrastive learning forces the network to impose a similar structure in the feature space by pulling the positive pairs closer to each other while keeping the negatives apart.

This binary separation into positives and negatives can be limiting whenever the boundary between those is blurry. For example, different samples from the same classes are used as negatives for \emph{instance recognition}, which prevents the network from exploiting their similarities. One way to address this issue is supervised contrastive learning (SCL)~\cite{khosla2020supervised}, which takes class labels into account when making pairs: samples from the same class are treated as positives, while samples of different classes pose negatives. However, even in this optimal setting with ground truth labels, the problem persists -- semantically similar classes share many visual features \cite{deselaers2011visual} with the query -- and some samples cannot clearly be categorized as either positive or negative, \eg the dog breeds in Fig.~\ref{fig:ranking}. Treating them as positives makes the network invariant towards the distinct attributes of the samples. As a result, the network struggles to distinguish between different dog breeds. If they are treated as negatives, the network cannot exploit their similarities. For transfer learning to other tasks, \eg out-of-distribution detection, a clean structure of the embedding space, s.t.~samples sharing certain attributes will be closer, is beneficial.

Another example comes from video representation learning: In addition to spatial crops as for images, videos allow to create temporal crops, \ie creating a sample from different frames of the same video. To date, it is an open point of discussion whether temporally different clips from the same video should be treated as positive \cite{feichtenhofer2021large} or negative \cite{dave2021tclr}. Treating them as positives will force the network to be invariant towards changes over time, but treating them as negatives will encourage the network to ignore the features that stay constant. In summary, a binary classification in positive and negative will, for most applications, lead to a sub-optimal solution. To the best of our knowledge, a method that benefits from a fine-grained definition of negatives, positives and various states in between is missing.

As a remedy, we propose Ranking Info Noise Contrastive Estimation (RINCE). RINCE supports a fine-grained definition of negatives and positives. Thus, methods trained with RINCE can take advantage of various kinds of similarity measures. For example similarity measures can be based on class similarities, gradual changes of content within videos, pretrained feature embeddings, or even the camera positions in a multi-view setting etc. In this work, we demonstrate class similarities and gradual changes in videos as examples.

RINCE puts higher emphasis on similarities between related samples than SCL and cross-entropy, resulting in a richer representation. We show that RINCE learns to represent semantic similarities in the embedding space, s.t.~more similar samples are closer than less similar samples. Key to this is a new \infoNCE-based loss, which enforces gradually decreasing similarity with increasing rank of the samples.

The representation learned with RINCE on \cifarhun improves significantly over cross-entropy for classification, retrieval and OOD detection, and outperforms the stronger SCL baseline~\cite{khosla2020supervised}. Here, improvements are particularly large for retrieval and OOD detection. To obtain ranked positives for RINCE, we use the superclasses of \cifarhun. Further, we demonstrate that RINCE works on large scale datasets and in more general applications, where ranking of samples is not initially given and contains noise. To this end, we show that RINCE outperforms our baselines on ImageNet-100 using only noisy ranks provided by an off-the-shelf natural language processing model \cite{liu2019roberta}. Finally, we showcase that RINCE can be applied to the fully unsupervised setting, by training RINCE unsupervised on videos, treating temporally far clips as weak positives. This results in a higher accuracy on the downstream task of video action classification than our baselines and even outperforms recent video representation learning methods.

In summary, our contributions are: 1) We propose a new \infoNCE-based loss that replaces the binary definition of positives and negatives by a ranked definition of similarity. 2) We study the properties of RINCE in a controlled supervised setting. Here, we show mild improvements on \cifarhun classification and sensible improvements for OOD detection. 3) We show that RINCE can handle significant noise in the similarity scores and leads to improvements on large scale datasets. 4) We demonstrate the applicability of RINCE to self-supervised learning with noisy similarities in a video representation learning task and show improvements over \infoNCE in all downstream tasks. 5) Code is available at\footnote{\url{https://github.com/boschresearch/rince}}.

\section{Related Works}

\paragraph{Contrastive Learning.}
Contrastive learning has recently advanced the field of self-supervised learning. Current state-of-the-art methods use \textit{instance recognition}, originally proposed by~\cite{Dosovitskiy:2016:DUF}, where the task is to recognize an instance under various transformations. Modern instance recognition methods utilize \textit{\infoNCE}~\cite{Oord:2018:CPC}, which was first proposed as \textit{N-pair} loss in~\cite{Sohn:2016:Npair}. It maximizes the similarity of \textit{positive pairs} -- which are obtained from two different views of the same instance -- while minimizing the similarity of \textit{negative pairs}, \ie views of different instances. Different views can be generated from multi-modal data \cite{Tian:2019:CMC}, permutations~\cite{Misra:2019:PIRL}, or augmentations~\cite{Chen:2020:SimCLR}. The negative pairs play a vital role in contrastive learning as they prevent shortcuts and collapsed solutions. In order to provide challenging negatives, \cite{He:2019:MoCo} introduce a memorybank with a momentum encoder, which allows to store a large set of negatives. Other approaches explicitly construct \textit{hard negatives} from patches in the same image~\cite{Oord:2018:CPC} or temporal negatives in videos~\cite{WACV_BFP}. More recent works omit negative pairs completely~\cite{Chen:2020:SimSiam,Grill:2020:BYOL}.

In the above cases, positive pairs are obtained from the same instance, and different instances serve as negatives even when they share the same semantics. Previous work addresses this issue by allowing multiple positive samples: \cite{Miech:2020:MILNCE} allows several positive candidates within a video, \cite{Han:2020:CoCLR} and \cite{Caron:2020:SwAV} obtain positives by clustering the feature space, whereas \cite{khosla2020supervised} uses class labels to define a set of positives. False negatives are eliminated from the \infoNCE loss by \cite{huynh2020boosting}, either using labels or a heuristic. Integrating multiple positives in contrastive learning is not straightforward: the set of positives can be noisy and include some samples that are more related than others.
In this work, we provide a tool to properly incorporate such samples.

\paragraph{Supervised Contrastive Learning.}
Labelled training data has been used in many recent works on contrastive learning.
\cite{Romijnders:2021:object_centric} use pseudo labels obtained from a detector, \cite{tian2020makes} use labels to construct better views and \cite{neill2021semantically} use similarity of class word embeddings to draw hard negatives.
The term \emph{supervised contrastive learning} (SCL) is introduced in \cite{khosla2020supervised} showing that SCL outperforms standard cross-entropy.

In the SCL setting ground truth labels are available and can be used to define positives and negatives. Commonly, samples from the same class are treated as positive, while instances from all other classes are treated as negatives. \cite{khosla2020supervised} find that the SCL loss function outperforms cross-entropy in the supervised setting. In contrast, \cite{huynh2020boosting} aim for an unsupervised detection of false negatives. They propose to only eliminate false negatives from the \infoNCE loss which leads to best results for noisy labels.

Along these lines, \cite{winkens2020contrastive} show that \infoNCE loss is better suited for out-of-distribution detection than cross-entropy. Here, we introduce a method to deal with non-binary similarity labels and study different versions of it in the SCL setting free from label noise and show that we get similar results in more noisy and even unsupervised settings.

\paragraph{Ranking.}
\textit{Learning to Rank} has been studied extensively~\cite{Burges:2005:L2RGD, cakir2019deep, Cao:2007:L2R, Liu:2009:L2R}. These works aim for downstream applications that require ranking \eg image or document retrieval, Natural Language Processing and Data Mining. In contrast, we are not interested in the ranking per-se, but rather use the ranking task to improve the learned representation. 

Some approaches in the field \emph{metric learning} use ranking losses to learn a feature embedding: Contrastive losses such as triplet loss~\cite{Weinberger:2006:triplet} or N-pair loss~\cite{Sohn:2016:Npair} can be interpreted as ranking the positive higher w.r.t. the anchor than the negative. For instance, \cite{Tschannen:2020:VIVI} use the triplet loss, to learn representations, but focus on learning invariances. \cite{ge2018deep} learn a hierarchy from data for hard example mining to improve the triplet loss. Further, these approaches only consider two ranks, whereas our method can work with multiple ranks.

\section{Methods}
\label{sec:methods}

\subsection{InfoNCE}
\label{sec:InfoNCE}
We start with the most basic form of the \textit{\infoNCE}. In this setting, two different views of the same data -- \eg two different augmentations of the same image -- are pulled together in feature space, while pushing views of different samples apart. More specifically, for a query $q$, a single positive $p$ and a set of negatives $\mathcal{N}=\{n_1,\dots n_k\}$ is given. The views are fed to an encoder network $f$, followed by a projection head $g$~\cite{Chen:2020:SimCLR}. To measure the similarity between a pair of features we use the cosine similarity $\cossim$. Overall the task is to train a critic $h(x,y)=\cossim\big(g(f(x)), g(f(y))\big)$ using the loss:
\begin{equation}
    \mathcal{L} = -\log \frac{\exp{\big(\frac{h(q, p)}{\tau}\big)}}{\exp{\big(\frac{h(q, p)}{\tau} \big)} + \sum\limits_{n\in\mathcal{N}}\exp{\big(\frac{h(q, n)}{\tau} \big)}},
    \label{eq:InfoNCE}
\end{equation}
where $\tau$ is a temperature parameter~\cite{Chen:2020:SimCLR}.
The above loss relies on the assumption that a single positive pair is available. One drawback with this approach is that all other samples are treated as negatives, even if they are semantically close to the query. Potential solutions include removing them from the negatives~\cite{Zhao:2021:WMI} or adding them to the positives~\cite{khosla2020supervised}, which we denote by $\mathcal{P} = \{p_1,\dots, p_l\}$. In other cases, we naturally have access to more than one positive, \eg we can sample several clips from a single video, see Fig.~\ref{fig:ranking_videos}. Having multiple positives per query leaves two options, which we discuss in the following.

\paragraph{Log$_{\text{out}}$ Positives.} 
A straightforward approach to include multiple positives is to compute Eq.~\eqref{eq:InfoNCE} for each of them, \ie take the sum over positives outside of the $\log$. This enforces similarity between all positives during training, which suits a clean set of positives well. 

\begin{equation}
    \mathcal{L}^{\text{out}} = -\sum\limits_{p \in \mathcal{P}}\log \frac{\exp{\big(\frac{h(q, p)}{\tau}\big)}}{\exp{\big(\frac{h(q, p)}{\tau}\big)} + \sum\limits_{n\in\mathcal{N}}\exp{\big(\frac{h(q, n)}{\tau}\big)}}.
    \label{eq:InfoNCE_out}
\end{equation}
However, the set of positives can be noisy, \eg sampling a temporally distant clip may include sub-optimal positives due to drastic changes in the video.

\paragraph{Log$_{\text{in}}$ Positives.} 
An alternative approach, which is more robust to noise or inaccurate samples \cite{Miech:2020:MILNCE}, is to take the sum inside the $\log$, Eq.~\eqref{eq:InfoNCE_in}. To minimize this loss, the network is not forced to set a high similarity to all pairs. It can neglect the noisy/false positives, given that a sufficiently large similarity is set for the true positives, see Tab.~\ref{table:temporal_ranking}. However, if a discrepancy between positives exists, it results in a degenerate solution of discarding hard positives. For instance, consider supervised learning where both augmentations and class positives are available for a given query: the class positives, which are harder to optimize, can be ignored.

\begin{equation}
    \mathcal{L}^{\text{in}} = -\log \frac{\sum\limits_{p \in \mathcal{P}}\exp{\big({\frac{h(q, p)}{\tau}}\big)}}{\sum\limits_{p \in \mathcal{P}}\exp{\big(\frac{h(q, p)}{\tau}\big)} + \sum\limits_{n\in\mathcal{N}}\exp{\big(\frac{h(q, n)}{\tau}}\big)}.
    \label{eq:InfoNCE_in}
\end{equation}

The above methods assume a binary set of positives and negatives. Thus, they can not exploit the similarity of positives and negatives. In the following, we discuss the proposed ranking version of \infoNCE that allows us to preserve the order of the positives and benefit from the additional information.

\subsection{RINCE: Ranking InfoNCE}
\label{sec:rince}

Let us assume that for a given query sample $q$, we have access to a set of ranked positives in a form of $\mathcal{P}_1,\dots,\mathcal{P}_r$, where $\mathcal{P}_i$ includes the positives of rank $i$. Let us also assume $\mathcal{N}$ is a set of negatives. Our objective is to train a critic $h$ such that:

\begin{equation}
h(q,p_1) >  \dots > h(q,p_r) > h(q,n) \quad \forall p_i \in \mathcal{P}_i, n\in \mathcal{N}.
\end{equation}
Note that $\mathcal{P}_i$ can contain multiple positives. For ease of notation we omit these indices. To impose the desired ranking presented by the positive sets, we use \infoNCE in a recursive manner where we start with the first set of positives, treat the remaining positives as negatives, drop the current positive, and move to the next. We repeat this procedure until there are no positives left. More precisely, the loss function reads $\mathcal{L}_{\text{rank}} = \sum_{i=1}^{r} \ell_i$, where
\begin{equation}
\ell_i = -\log \frac{\sum\limits_{p \in \mathcal{P}_i} \exp{\big(\frac{h(q,p)}{\tau_i}\big)}}{\sum\limits_{p \in \bigcup_{j\geq i} \mathcal{P}_j} \exp{\big(\frac{h(q,p)}{\tau_i}\big)} + \sum\limits_{n \in \mathcal{N}} \exp{\big(\frac{h(q,n)}{\tau_i}}\big)}
\label{eq:ithrank}
\end{equation}
and $\tau_i < \tau_{i+1}$. Eq.~\eqref{eq:ithrank} denotes the $\mathcal{L}^{\text{in}}$ version of \infoNCE for positives of same rank; other variants are summarized in Tab.~\ref{tab:losses}.
The rational behind this loss is simple: The $i$-th loss is optimized when I) $\exp(h(q,p_i)/\tau_i) \gg 0$, II) $\exp(h(q,p_j)/\tau_i) \to 0$ for $j > i$ and III) $\exp(h(q,n)/\tau_i) \to 0 \text{ for all } i,j,n$. I) and II) are competing across the losses: $\ell_i$ entails $\exp(h(q,p_{i+1})/\tau_i) \to 0$ but $\ell_{i+1}$ requires $\exp(h(q,p_{i+1})/\tau_{i+1}) \gg 0$. This requires the model to trade-off the respective loss terms, resulting in a ranking of positives $h(q,p_{i}) > h(q,p_{i+1})$. 

In the following we explain the intuition behind our choice of $\tau$ values based on the analyses of \cite{wang2021understanding}; for a more detailed analysis see \supmat A low temperature in the \infoNCE loss results in a larger relative penalty on the high similarity regions, \ie hard negatives. As the temperature increases, the relative penalty distributes more uniformly, penalizing all negatives equally. A low temperature in $\ell_i$ allows the network to concentrate on forcing $h(q,p_i) > h(q,p_{i+1})$, ignoring easy negatives. A higher temperature on $\ell_r$ relaxes the relative penalty of negatives with respect to $p_r$ so that the network can enforce $h(q,p_r) > h(q,n)$.

\begin{table}[h]
    \centering
    \small
    \begin{tabular}{l|ll}
        Naming & \# positives per rank & loss \\
        \midrule
        \RINCEsingle & single & Eq.~\eqref{eq:InfoNCE}\\
        \RINCEout & multiple & Eq.~\eqref{eq:InfoNCE_out} \\
        \RINCEin & multiple & Eq.~\eqref{eq:InfoNCE_in} \\
        \multirow{2}{*}{\RINCEoutin} & \multirow{2}{*}{multiple} & Eq.~\eqref{eq:InfoNCE_out} ($\ell_1$);\\ & & Eq.~\eqref{eq:InfoNCE_in} ($\ell_i, i>1$) \\
    \end{tabular}
    \caption{{\bf Different variants of RINCE.} 
    For the exact loss functions see the \supmat}
    \label{tab:losses}
\end{table}

\section{Experiments}

We first study the properties of RINCE in the controlled supervised setting, looking at classification accuracy, retrieval and out-of-distribution (OOD) detection on \cifarhun. Next, we show that RINCE leads to significant improvements on the large scale dataset ImageNet-100 in terms of accuracy and OOD, even with more noisy similarity scores.
Last, we showcase exemplary with unsupervised video representation learning that RINCE can be used in an unsupervised setting.
For all experiments we follow the MoCo v2 setting~\cite{chen2020improved} with a momentum encoder, a memory bank and a projection head.
Throughout the section we compare different versions of RINCE (Tab.~\ref{tab:losses}), to study their behavior in different settings. 
More ablations in the \supmat

\subsection{Learning from Class Hierarchies}
\label{sec:supCon}

The optimal testbed to study the proposed loss functions is the supervised contrastive learning (SCL) setting.
The effect of the proposed loss functions can be studied without confounding noise, using ground truth labels and ground truth rankings.
In SCL all samples with the same class 
are considered as positives, thus either Eq.~\eqref{eq:InfoNCE_out}, or Eq.~\eqref{eq:InfoNCE_in} is used.
However, semantically similar classes share similar visual features \cite{deselaers2011visual}. 
When strictly treated as negatives the model does not mirror the structure available by the labels in its feature space. This, however, is favorable for transferability to other tasks.
RINCE allows the model to keep this structure, and learn not only dissimilarities between, but also similarities across classes.
We show quantitatively that RINCE learns a higher quality representation than cross-entropy and SCL on \cifarhun and ImageNet-100 by evaluating on linear classification, image retrieval, and OOD tasks.
Unless otherwise stated, we report results for ResNet-50. 
More implementation details in the \supmat

\paragraph{Datasets.}
\cifarhun \cite{krizhevsky2009learning} provides both, class and superclass labels, defining a semantic hierarchy.
We use this hierarchy to define first rank positives (same class) and second rank positives (same superclass).

\tinyimagenet \cite{le2015tiny} comprises 200 ImageNet \cite{deng2009imagenet} classes at low resolution.
ImageNet-100 \cite{Tian:2019:CMC} is a 100 class subset of ImageNet.
We use the RoBERTa \cite{liu2019roberta} model to obtain semantic word embeddings for all class names.
Second rank positives are based on the word embedding similarity and a predefined threshold. 
Details in the \supmat

\begin{table*}[h]
\small
    \centering
    \begin{tabular}{lccc|cc}
        \toprule
        \multirow{2}{*}{Method}  & \multicolumn{2}{c}{Cifar100 fine} & Cifar100 superclass & \multicolumn{2}{c}{AUROC} \\
         & Accuracy & R@1 & R@1 & $\mathcal{D}_{\text{out}}$: \cifarten & $\mathcal{D}_{\text{out}}$: \tinyimagenet \\
        \midrule
         SCL-out & 76.50 & N/A & N/A & N/A & N/A \\
        Soft Labels$^\circ$  &  76.90 & N/A & N/A & N/A & 67.50\\
        ODIN$^\dagger$  & N/A & N/A & N/A & 77.20 & 85.20 \\
         Mahalanobis$^\dagger$  & N/A & N/A & N/A & 77.50 & 97.40 \\
         Contrastive OOD$^\ddag$  & N/A & N/A & N/A  & 78.30 & N/A\\
         Gram Matrices & N/A & N/A & N/A  &  67.90 & 98.90 \\
        \midrule
        Cross-entropy$^*$ &   74.52 $\pm$ 0.32 & 74.84 $\pm$ 0.21 & 83.99 $\pm$ 0.21 & 75.32 $\pm$ 0.65 & 77.76 $\pm$ 0.77 \\
        Cross-entropy s.a.$^*$  & 75.46 $\pm$ 1.09 & 76.03 $\pm$ 1.04 & 84.68 $\pm$ 0.86 & 75.91 $\pm$ 0.10 & 79.44 $\pm$ 0.50 \\
        Triplet & 68.44 $\pm$ 0.18 & 47.73 $\pm$ 0.14 & 72.29 $\pm$ 0.27  & 70.33 $\pm$ 0.54 & 80.76 $\pm$ 0.24\\
        Hierarchical Triplet$^*$ & 69.27 $\pm$ 1.64 & 65.31 $\pm$ 2.69 & 77.41 $\pm$ 1.55 & 71.97 $\pm$ 2.48 & 76.22 $\pm$ 1.27\\
        Fast AP$^*$ & 66.96 $\pm$ 0.88 & 62.03 $\pm$ 0.51 & 69.56 $\pm$ 0.54 & 69.14 $\pm$ 1.02 & 72.44 $\pm$ 0.94\\
        Smooth Labels &75.66 $\pm$ 0.27 & 74.90 $\pm$ 0.06 &85.59 $\pm$ 0.12 & 74.35 $\pm$ 0.65 & 80.10 $\pm$ 0.77 \\
        Two heads &  74.08 $\pm$ 0.40 & 73.62 $\pm$ 0.31 & 81.92 $\pm$ 0.21 & \textbf{77.99 $\pm$ 0.07} & 78.35 $\pm$ 0.39 \\
        SCL-in superclass$^*$  & 74.41 $\pm$ 0.15 & 69.83 $\pm$ 0.28 & 85.35 $\pm$ 0.51 & 74.40 $\pm$ 0.72 & 80.20 $\pm$ 1.05 \\
        SCL-in$^*$  & 76.86 $\pm$ 0.18 & 73.20 $\pm$ 0.19 & 82.16 $\pm$ 0.24 & 74.63 $\pm$ 0.16 & 78.96 $\pm$ 0.45 \\
        SCL-out$^*$  & 76.70 $\pm$ 0.29 & 74.45 $\pm$ 0.39 & 82.94 $\pm$ 0.39 & 75.32 $\pm$ 0.59 & 79.80 $\pm$ 0.70 \\
        SCL-in two heads$^*$  & 77.15 $\pm$ 0.14 & 74.36 $\pm$ 0.10 & 83.31 $\pm$ 0.09 & 75.41 $\pm$ 0.16 & 79.34 $\pm$ 0.19 \\
        SCL-out two heads$^*$  & 76.91 $\pm$ 0.08 & 74.87 $\pm$ 0.37 & 83.74 $\pm$ 0.16 & 75.27 $\pm$ 0.34 & 79.64 $\pm$ 0.53 \\
        Contrastive OOD & N/A & N/A & N/A & 74.20 $\pm$ 0.40 & N/A\\
        \RINCEout &  76.94  $\pm$ 0.16 & 76.68 $\pm$ 0.09 & 86.10 $\pm$ 0.25 & \underline{77.76 $\pm$ 0.09} & 81.02 $\pm$ 0.14 \\
        \RINCEoutin & \textbf{77.59 $\pm$ 0.21} &\underline{77.47 $\pm$ 0.16} & \underline{86.20 $\pm$ 0.23} & 76.82 $\pm$ 0.44 & \underline{81.40 $\pm$ 0.38} \\
        \RINCEin &  \underline{77.45 $\pm$ 0.05} & \textbf{77.56 $\pm$ 0.03} & \textbf{86.46 $\pm$ 0.21} & 77.03 $\pm$ 0.53 & \textbf{81.78 $\pm$ 0.05} \\
        \bottomrule
    \end{tabular}
    \caption{\textbf{Classification, retrieval and OOD results for \cifarhun pretraining.}
    Left: classification and retrieval; fine-grained task (fine) with 100 classes and superclass task (superclass) with 20 classes.
    Right: OOD task with inlier dataset $\mathcal{D}_\text{in}$: \cifarhun and outlier dataset $\mathcal{D}_\text{out}$: \cifarten and \tinyimagenet.
    We report the mean and standard deviation over 3 runs. Contrastive OOD averaged over 5 runs.
    Best method in bold, second best underlined.
    Note that, models indicated with $^\dagger$ are not directly comparable, since they use data explicitly labeled as OOD samples for tuning. $^*$ indicates methods of others trained by us, $^\circ$ uses 2$\times$ wider ResNet-40, $^\ddag$ 4$\times$ wider ResNet-50. The lower part of the table uses ResNet-50. Methods not references in text: Soft Labels \cite{lee2020soft}, Gram Matrices \cite{sastry2020detecting}, Triplet \cite{Weinberger:2006:triplet}.
    }
\label{tab:both}
\end{table*}

\paragraph{Baselines and SOTA.} 
As baselines we use \textbf{cross-entropy}, cross-entropy with the same augmentations as RINCE (\textbf{cross-entropy s.a.}), \textbf{Triplet} loss \cite{Weinberger:2006:triplet} and \textbf{SCL} \cite{khosla2020supervised}, trained with Eq.~\eqref{eq:InfoNCE_out} ({\bf SCL-out}) or Eq.~\eqref{eq:InfoNCE_in} ({\bf SCL-in}).
An advantage of RINCE compared to these baselines is that it benefits from extra information provided by the superclasses. 
To show that making use of this knowledge is not trivial, we compare to the following baselines: 1) We train SCL on \cifarhun with 20 superclasses, denoted by \textbf{SCL superclass}. 
2) \textbf{Hierarchical Triplet} \cite{ge2018deep}, which uses the superclasses to mine hard examples.
3) \textbf{Fast AP} \cite{cakir2019deep}, a ``learning to rank'' approach that directly optimizes Average Precision.
4) \textbf{Label smoothing} \cite{szegedy2016rethinking}, which reduces network over-confidence and can improve OOD detection \cite{lee2020soft}.
We assign some probability mass to the classes from the same superclass.
5) A multi-classification baseline, referred to as \textbf{two heads}, that jointly predicts class and superclass labels.
6) \textbf{SCL two heads}, a variant of two heads, that uses the SCL loss instead of cross-entropy.
Details for all baselines are given in the \supmat

\paragraph{Classification and Retrieval on Cifar.}

For the classification evaluation we train a linear layer on top of the last layer of the frozen pre-trained networks. The non-parametric retrieval evaluation involves finding the relevant data points in the feature space of the pre-trained network in terms of class labels via a simple similarity metric, \eg cosine similarity. RINCE is superior to the baselines for all experiments, Tab.~\ref{tab:both}.
Note, that all evaluations in Tab.~\ref{tab:both} are based on the same pre-trained weights using \cifarhun fine labels as rank 1 and, if applicable, superclass labels as rank 2.

\begin{figure*}[ht!]
    \centering
    \begin{subfigure}{.33\textwidth}
  \centering
  \includegraphics[width=1\linewidth]{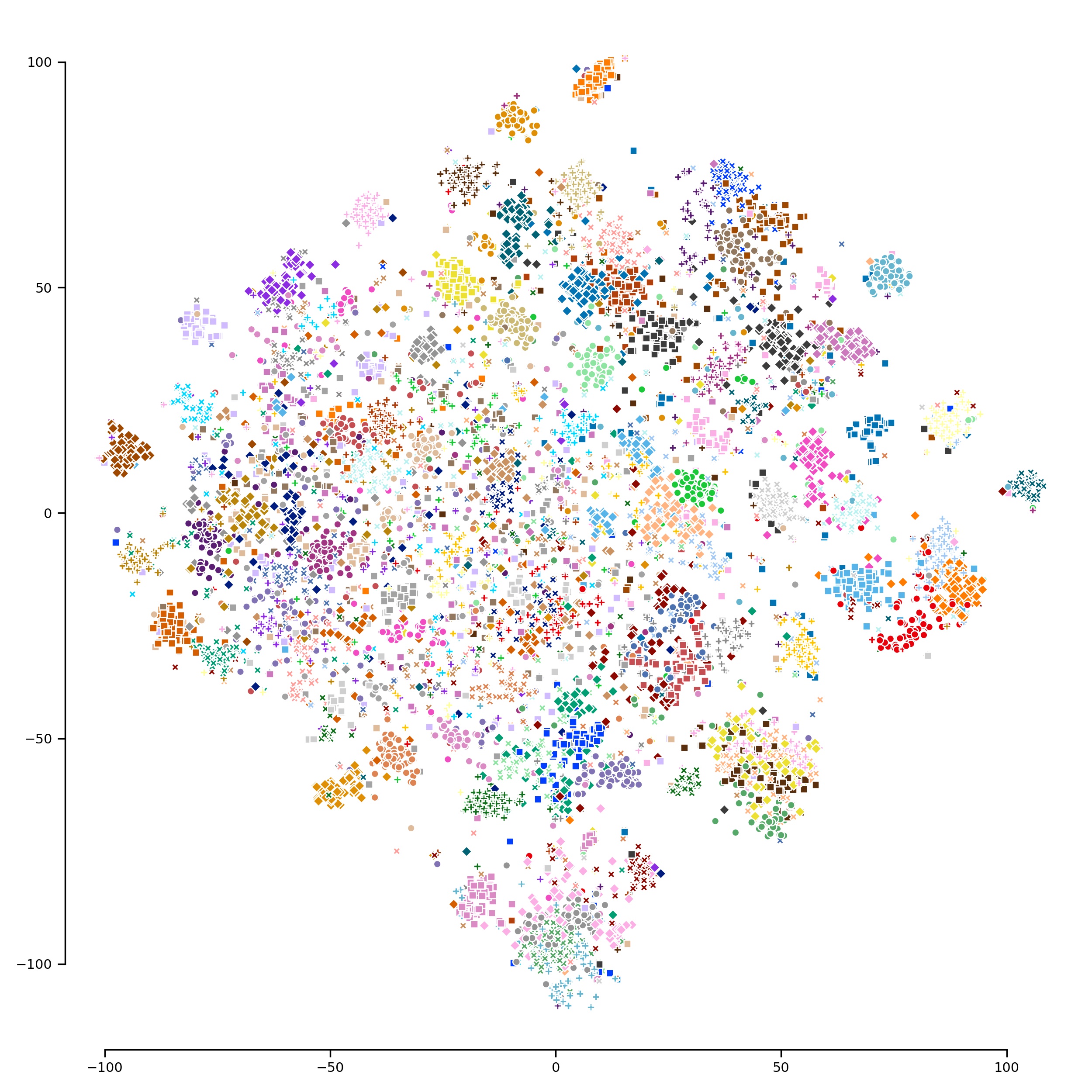}
  \caption{}
  \label{fig:tsne_SCL}
    \end{subfigure}%
    \begin{subfigure}{.33\textwidth}
  \centering
      \includegraphics[width=1\linewidth]{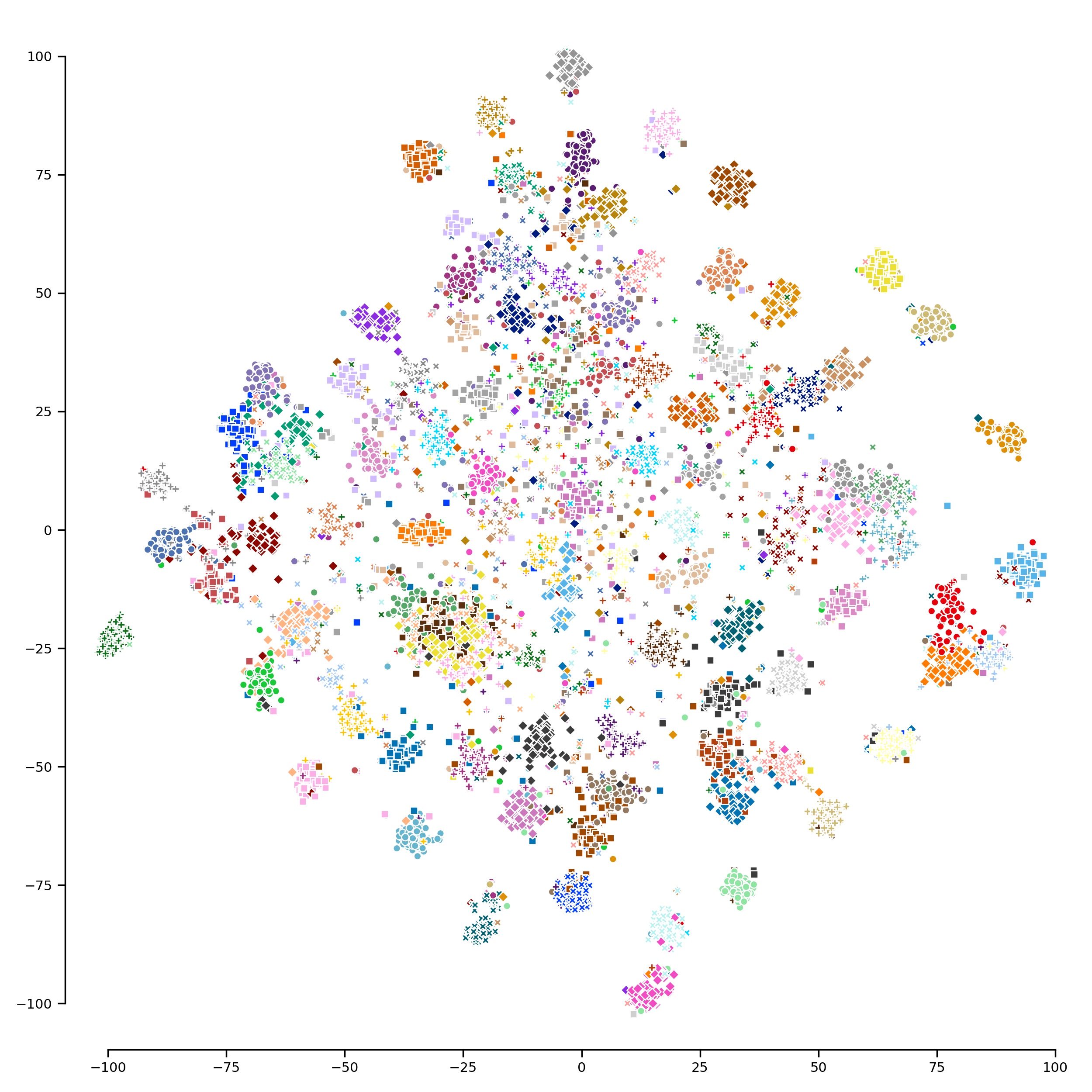}
    \caption{}
    \label{fig:RINCE_in}

\end{subfigure}
\begin{subfigure}{.33\textwidth}
    \centering
    \includegraphics[width=1\linewidth]{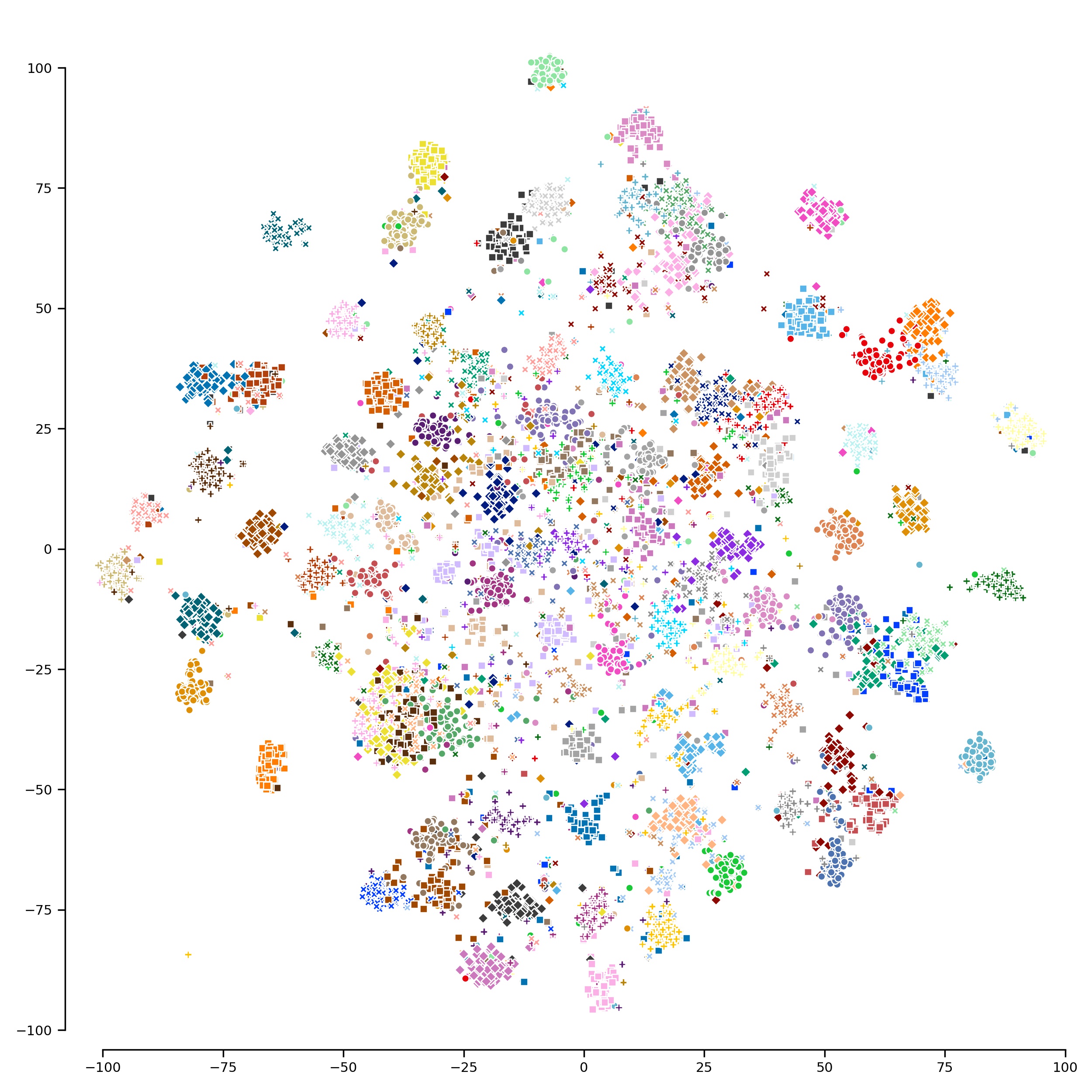}
    \caption{}
    \label{fig:tsne_RINCE}
    \end{subfigure}
\caption{
{\bf Qualitative comparison of embedding spaces.}
T-SNE plot of (a) supervised contrastive learning (\emph{SCL-in}) and (b) \emph{\RINCEin} (c) \emph{\RINCEoutin} on \cifarhun. Best seen in color, on screen and zoomed in. Color and marker type combined indicate class. Labels omitted for clarity. \supmat contains a version of this plot with color indicating the superclass.
RINCE learns a more structured embedding space than SCL, e.g.~classes are linearly separable and can be modelled well by a Gaussian.
}
\label{fig:tsne}
\end{figure*}

These experiments indicate that training with RINCE maintains ranking order and results in a more structured feature space in which the samples of the same class are well separated from the other classes. This is further approved by a qualitative comparison between embedding spaces in Fig.~\ref{fig:tsne}.

Furthermore, we find that the grouping of classes is learned by the MLP head. The increased difficulty of the ranking task of RINCE results in a more structured embedding space before the MLP compared with SCL, see \supmat Fig.~\ref{fig:confusion_mat}.

\paragraph{Out-of-distribution Detection.}

To further investigate the structure of the learned representation of RINCE 
we evaluate on the task of out-of-distribution detection (OOD). 
As argued in \cite{winkens2020contrastive}, 
models trained with cross-entropy only need to distinguish classes and can omit irrelevant features.
Contrastive learning differs, by forcing the network to distinguish between each pair of samples, resulting in a more complete representation. Such a representation is beneficial for OOD detection \cite{hendrycks2019using,winkens2020contrastive}. 
Therefore, OOD performance can be seen as evaluation of representation quality beyond standard metrics like accuracy and retrieval.
RINCE  incentivizes the network to learn an even richer representation.
Besides that, OOD benefits from good trade-off between alignment and uniformity, which RINCE manages well (Fig.~\ref{fig:unif_align_full} in \supmat). 

We follow common evaluation settings for OOD \cite{lee2018simple,liang2018enhancing,winkens2020contrastive}. Here \cifarhun is used as the inlier dataset $\mathcal{D}_\text{in}$, \cifarten and \tinyimagenet as outlier dataset $\mathcal{D}_\text{out}$.
Note that \cifarhun and \cifarten have disjoint labels and images. 
For both protocols we only use the test or validation images. 
Our models are identical to those in the previous section.
Inspired by \cite{winkens2020contrastive}, we follow a simple approach, and fit class-conditional multivariate Gaussians to the embedding of the training set. We use the log-likelihood to define the OOD-score. 
As a result, the likelihood to identify OOD-samples is high, if each in-class follows roughly a Gaussian distribution in the embedding space, compare Fig.~\ref{fig:tsne_SCL} and \ref{fig:tsne_RINCE}.
For evaluation, we compute the area under the receiver operating characteristic curve (AUROC), details in \supmat

Results and a comparison to the most related previous work is shown in Tab.~\ref{tab:both}. 
Note that we aim here to compare the learned representation space via RINCE to its counterparts, \ie cross-entropy and SCL, but show well known methods as reference.
Most importantly, RINCE clearly outperforms cross-entropy, all SCL variants, contrastive OOD and our own baselines using the identical OOD approach. 
Only, two-heads outperforms all other methods in the near OOD setting with $\mathcal{D}_\text{out}$: Cifar10. However, performance on all other settings is low, showing weak generalization.
This underlines our hypothesis, that training with RINCE yields a more structured and general representation space.
Comparing to related works, RINCE not only outperforms Contrastive OOD~\cite{winkens2020contrastive} using the same architecture, but even approaches the $4\times$ wider ResNet on Cifar10 as $\mathcal{D}_\text{out}$. 
ODIN~\cite{liang2018enhancing} and Mahalanobis~\cite{lee2018simple} require samples labelled as OOD to tune parameters of the OOD approach. Here we evaluate in the more realistic setting without labelled OOD samples. Despite using significantly less information, RINCE is compatible with them and even outperforms them for $\mathcal{D}_\text{out}$: Cifar10.

\begin{table}[t]
    \small
    \centering
    \begin{tabular}{lc|cc}
        \toprule
        \multirow{3}{*}{Method} & \multicolumn{1}{c|}{ } & \multicolumn{2}{c}{AUROC }\\
        & \multirow{2}{*}{Accuracy} & $\mathcal{D}_{\text{out}}$: &  $\mathcal{D}_{\text{out}}$:  \\
        & & ImageNet-100$^\dagger$ & AwA2\\
        \midrule
        Cross-entropy s.a. & 83.94 & 79.076 $\pm$ 1.477 & 79.04 \\
        SCL-out & 84.18 & 79.779 $\pm$ 1.274 &79.05  \\
        \RINCEoutin  &\textbf{84.90} & \textbf{80.473 $\pm$ 1.210} & \textbf{80.73}\\
        \bottomrule
    \end{tabular}
        \caption{\textbf{ImageNet-100 classification accuracy and OOD detection for $\mathcal{D}_{\text{in}}$: ImageNet-100, and $\mathcal{D}_{\text{out}}$: ImageNet-100$^\dagger$ and AwA2~\cite{Xian:2018:AwA2}.}
        ImageNet-100$^\dagger$ denotes three ImageNet-100 datasets with non-overlapping classes.}
    \label{tab:imagenet}
\end{table}

\subsection{Large Scale Data and Noisy Similarities}

Additionally, we perform  the same evaluations on ImageNet-100, a 100-class subset of ImageNet, see Tab.~\ref{tab:imagenet}. Here, we use ResNet-18. 
We obtain the second rank classes for a given class via similarities of the RoBERTa \cite{liu2019roberta} class name embeddings. In contrast to the previous experiments, where ground truth hierarchies are known, these similarity scores are noisy and inaccurate -- yet it still provides valuable information to the model.
We evaluate our model via linear classification on ImageNet-100 and two OOD tasks: 
AwA2~\cite{Xian:2018:AwA2} as $\mathcal{D}_{\text{out}}$ and ImageNet-100$^{\dagger}$, where we use the remaining ImageNet classes to define three non-overlapping splits and report the average OOD.

Result are shown in Tab.~\ref{tab:imagenet}. Again, RINCE significantly improves over SCL and cross-entropy in linear evaluation as well as on the OOD tasks. 
This demonstrates 1) that RINCE can handle noisy rankings and 2) that RINCE leads to improvements on large scale datasets.
Next, we move to an even less controlled setting and define a ranking based on temporal ordering for unsupervised video representation learning.

\begin{figure*}[h!]
    \centering
    \includegraphics[width=\linewidth]{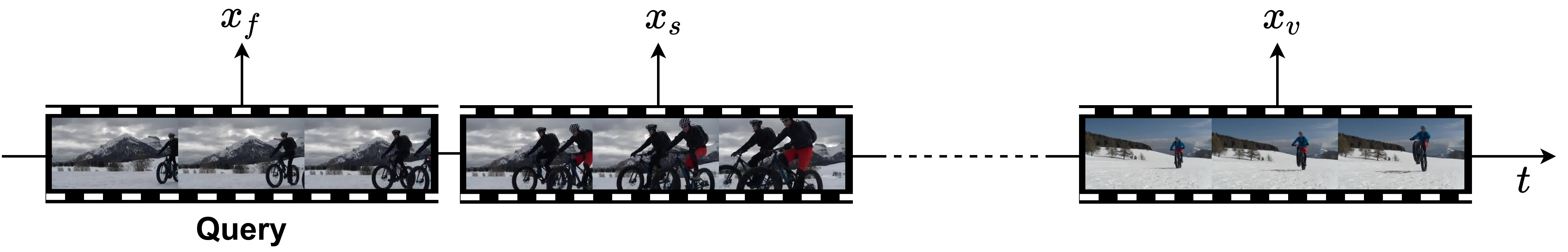}
    \caption{\textbf{Positives in Videos.} For a given query clip we use \textit{frame} positives $x_f$, \textit{shot} positives $x_s$ and \textit{video} positives $x_v$.}
    \label{fig:ranking_videos}
\end{figure*}
\subsection{Unsupervised RINCE}
\label{sec:unsup}

\begin{table*}[h!]
\small
\centering
    \begin{tabular}{llcc|cc|cc}
    \toprule
    \multirow{2}{*}{Method} & \multirow{2}{*}{Loss} & \multirow{2}{*}{Positives} & \multirow{2}{*}{Negatives} & \multicolumn{2}{c|}{Top 1 Accuracy} & \multicolumn{2}{c}{Retrieval mAP}\\
    & & & & HMDB & UCF & HMDB & UCF \\
    \midrule
    VIE  & - & - & - & $44.8$ & $72.3$ & - & - \\
    LA-IDT  & - & - & - & $44.0$ & $72.8$ & - & - \\
    \midrule
    \infoNCE & $\mathcal{L}$ & $\{x_f\}$ & $\mathcal{N}$ & $41.5$ & $71.3$ & $0.0500$ & $0.0688$\\
    \multirow{2}{*}{hard positive} &$\mathcal{L}^{\text{in}}$ & $\{x_f,x_s,x_v\}$ & $\mathcal{N}$ & $42.6$ & $74.3$ & $0.0685$ & $0.1119$ \\
     &$\mathcal{L}^{\text{out}}$ & $\{x_f,x_s,x_v\}$ & $\mathcal{N}$ & $41.4$ & $73.6$ & $0.0666$ & $0.1204$ \\
     \multirow{2}{*}{easy positive} &$\mathcal{L}^{\text{in}}$ & $\{x_f,x_s\}$ & $\mathcal{N}$ & $42.7$ & $74.5$ & $0.0581$ & $0.1257$ \\
     &$\mathcal{L}^{\text{out}}$ & $\{x_f,x_s\}$ & $\mathcal{N}$ & $40.7$ & $73.5$ & $0.0593$ & $0.1297$ \\
    \multirow{2}{*}{hard negative} &$\mathcal{L}^{\text{in}}$ & $\{x_f,x_s\}$ & $\{x_v\} \cup \mathcal{N}$ & $43.6$ & $74.3$ & $0.0678$ & $0.1141$ \\
    &$\mathcal{L}^{\text{out}}$ & $\{x_f,x_s\}$ & $\{x_v\} \cup \mathcal{N}$ & $43.5$ & $75.2$ & $0.0675$ & $0.1193$ \\
    RINCE &\RINCEsingle & $x_f>x_s>x_v$ & $\mathcal{N}$ & $\mathbf{44.9}$ & $\mathbf{75.4}$ & $\mathbf{0.0719}$ & $\mathbf{0.1395}$ \\
    \bottomrule
    \end{tabular}
    \caption{
{\bf Finetuning on UCF and HMDB.} $\mathcal{L}$, $\mathcal{L}^{\text{in}}$ and $\mathcal{L}^{\text{out}}$ correspond to Eq.~\eqref{eq:InfoNCE}, Eq.~\eqref{eq:InfoNCE_in} and Eq.~\eqref{eq:InfoNCE_out}, respectively. \textit{Positives} and \textit{Negatives} indicates how $x_f, x_s, x_v$ were incorporated into contrastive learning, where $\mathcal{N}$ denotes the set of negative pairs from random clips. Since we consider only a single positive per rank we use the \RINCEsingle loss variant for RINCE.
} 
\label{table:temporal_ranking}
\end{table*}{}

In this section we demonstrate that RINCE can be used in a fully unsupervised setting with noisy hierarchies by applying it to unsupervised video representation. Inspired by \cite{Tschannen:2020:VIVI}, we construct three ranks for a given query video, same frames, same shot and same video, see Fig.~\ref{fig:ranking_videos}.

The first positive $x_f$ is obtained by augmenting the query frames. The second positive $x_s$ is a clip consecutive to the query frames, where small transformations of the objects, illumination changes, etc. occur. The third positive $x_v$ is sampled from a different time interval of the same video, which may show visually distinct but semantically related scenes. Naturally, $x_f$ shows the most similar content to the query frames, followed by $x_s$ and finally $x_v$. We compare temporal ranking with RINCE to different baselines.

\paragraph{Baselines.}
We compare to the basic \textbf{\infoNCE}, where a single positive is generated via augmentations \cite{Chen:2020:SimCLR, He:2019:MoCo}, \ie only \textit{frame} positives $x_f$. When considering multiple clips from the same video such as $x_s$ and $x_v$, there are several possibilities: We can treat them all as positives (\textbf{hard positive}), we can use the distant $x_v$ as a \textbf{hard negative} or ignore it (\textbf{easy positive}). In both cases $\mathcal{L}^{\text{out}}$, Eq.~\eqref{eq:InfoNCE_out}, and $\mathcal{L}^{\text{in}}$, Eq.~\eqref{eq:InfoNCE_in}, are possible. Additionally, we compare to two recent methods trained in comparable settings, \ie VIE \cite{Zhuang:2020:VIE}, LA-IDT \cite{Tokmakov:2020:LA_IDT}.

\paragraph{Ranking Frame-, Shot- and Video-level Positives.}
We sample short clips of a video, each consisting of $16$ frames. We augment each clip with a set of standard video augmentations. For more details we refer to the \supmat For the anchor clip $x$, we define positives as in Fig.~\ref{fig:ranking_videos}: $p_1=x_f$ consists of the same frames as $x$, $p_2=x_s$ is a sequence of $16$ frames adjacent to $x$, and $p_3=x_v$ is sampled from a different time interval than $x_f$ and $x_s$. Negatives $x_n$ are sampled from different videos. Since each rank $i$ contains only a single positive $p_i$, Eq.~\eqref{eq:InfoNCE_out} = Eq.~\eqref{eq:InfoNCE_in}, we call this variant \RINCEsingle. By ranking the positives we ensure that the similarities satisfy $\text{sim}(x,x_f)>\text{sim}(x,x_s)>\text{sim}(x,x_v)>\text{sim}(x,x_n)$, adhering to the temporal structure in videos.

\paragraph{Datasets and Evaluation.}
For self-supervised learning, we use Kinetics-400 \cite{Kay:2017:Kinetics} and discard the labels. Our version of the dataset consists of $234.584$ training videos. We evaluate the learned representation via finetuning on UCF~\cite{Soomro:2012:UCF} and HMDB~\cite{Kuehne:2011:HMDB} and report top 1 accuracy. In this evaluation, the pretrained weights are used to initialize a network and train it end-to-end using cross-entropy. Additionally, we evaluate the representation via nearest neighbor retrieval and report mAP. Precision-Recall curves can be found in the \supmat

\paragraph{Experimental Results.}
For all experiments we use a 3D-ResNet-18 backbone. Training details can be found in the \supmat We report the results for RINCE as well as the baselines in Tab.~\ref{table:temporal_ranking}. Adding shot- and video-level samples to \infoNCE improves the downstream accuracies. We observe that adding $x_v$ to the set of negatives to provide a hard negative rather than adding it to the set of positives leads to higher performance, suggesting that this should not be a true positive. This is further supported by the second and third row, where all three positives are treated as true positives. Here, $\mathcal{L}^{\text{out}}$, which forces all positives to be similar, leads to inferior performance compared to $\mathcal{L}^{\text{in}}$. $\mathcal{L}^{\text{in}}$ allows more noise in the set of positives by weak influence of false positives $x_v$. With RINCE we can impose the temporal ordering $x_f>x_s>x_v$ and treat $x_v$ properly, leading to the highest downstream performance. Improvements of RINCE over $\mathcal{L}^{\text{out}}$ is less pronounced on UCF. This is due to the strong static bias~\cite{Li:2018:RESOUND} of UCF and $\mathcal{L}^{\text{out}}$ encourages static features. Contrarily, improvements of RINCE over $\mathcal{L}^{\text{out}}$ on HMDB are substantial, due to the weaker bias towards static features. Last, we compare our method to two recent unsupervised video representation learning methods that use the same backbone network in Tab.~\ref{table:temporal_ranking}. We outperform these methods on both datasets.

\section{Conclusion}
\label{sec:conclusion}
We introduced RINCE, a new member in the family of \infoNCE losses. We show that RINCE can exploit rankings to learn a more structured feature space with desired properties, lacking with standard \infoNCE. Furthermore, representations learned through RINCE can improve accuracy, retrieval and OOD. Most importantly, we show that RINCE works well with noisy similarities, is applicable to large scale datasets and to unsupervised training. We compare the different variants of RINCE. Here lies a limitation: Different variants are optimal for different tasks and must be chosen based on domain knowledge. Future work will explore further applications of obtaining similarity scores, \eg based on distance in a pretrained embedding space, distance between cameras in a multi-view setting or distances between clusters.

\section*{Acknowledgments}

JG has been supported by the Deutsche Forschungsgemeinschaft (DFG, German Research Foundation) - GA1927/4-2.

\bibliography{biblio.bib}

\setcounter{equation}{5}
\setcounter{table}{4}
\setcounter{figure}{3}

\clearpage
\appendix

\section{Appendix}

\subsection{RINCE Loss Analysis}
In the following we will give a more theoretical analysis of RINCE, explain why it leads to ranking and justify our choice of setting $\tau_i < \tau_{i+1}$. First, we will study the relative penalty \cite{wang2021understanding} to identify which negatives contribute most to the individual RINCE terms, dependent on the choice of $\tau$ and elaborate how this results in ranking. Next, we will elaborate how the choice of $\tau_i < \tau_{i+1}$ guides the network towards a desired trade-off between the opposing terms in the RINCE loss.

\paragraph{Which loss term focuses on which negatives?}
Following \cite{wang2021understanding}, we investigate the impact of negatives in the loss relative to the impact of positives, which is referred to as \textit{relative penalty}. The relative penalty $r_n^p$ is obtained by dividing the gradient magnitude with respect to the similarity of negative and query ($s_{q,n}=h(q,n)$) by the gradient magnitude with respect to the similarity of positive and query ($s_{q,p}=h(q,p)$):
\begin{equation}
r_n^p = \left|\frac{\partial \ell(q)}{\partial s_{q,n}}\right|/\left|\frac{\partial \ell(q)}{\partial s_{q,p}}\right|.
\end{equation}
\textit{A large relative penalty for a given negative $n$ implies a large contribution of this negative in the \infoNCE loss}.
For simplicity, we consider only two ranks but extending the explanation to more ranks is trivial. Let $p_1\in\mathcal{P}_1$ denote a first rank positive, $p_2\in\mathcal{P}_2$ a second rank positive, $n\in\mathcal{N}$ a negative and capital letters denote the entire set. The relative penalty of a negative $n\in \mathcal{N}$ with respect to $p_2$ in $\ell_2$ is given by:
\begin{equation}
r_n^{p_2} = \frac{\exp(h(q,n)/\tau_2)}{\sum\limits_{x \in \mathcal{N}\setminus \{n\}} \exp(h(q,x)/\tau_2)}
\label{eq:rel_penalty_n_p2}
\end{equation}
Similarly, the relative penalty of $n$ with respect to $p_1$ in $\ell_1$ is:
\begin{equation}
r_n^{p_1} = \frac{\exp(h(q,n)/\tau_1)}{\sum\limits_{x \in \mathcal{N}\setminus \{n\}} \exp(h(q,x)/\tau_1) + \sum\limits_{p_2^* \in \mathcal{P}_2} \exp(h(q,p_2^*)/\tau_1)}\\
\label{eq:rel_penalty_n_p1}
\end{equation}
and for $p_2$ with respect to $p_1$ in $\ell_1$:
\begin{equation}
r_{p_2}^{p_1} = \frac{\exp(h(q,p_2)/\tau_1)}{\sum\limits_{x \in \mathcal{N}} \exp(h(q,x)/\tau_1) + \sum\limits_{p_2^* \in \mathcal{P}_2\setminus \{p_2\}} \exp(h(q,p_2^*)/\tau_1)}.
\label{eq:rel_penalty_p2_p1}
\end{equation}
Note that $p_2$ serves as a negative for $p_1$ in $\ell_1$ and a positive in $\ell_2$. With small $\tau_1$ Eq.~\eqref{eq:rel_penalty_n_p1} is larger for close samples of $n$ than with larger $\tau_1$. Therefore, small $\tau_1$ result in significantly larger relative penalties for $n$ close to $q$, in comparison to a larger $\tau_1$. Both also hold for Eq.~\eqref{eq:rel_penalty_p2_p1} and $p_2$. Thus, increasing $\tau$ shifts the focus of the loss function from close negatives to a more uniform contribution of all negatives. With $\tau_2>\tau_1$, Eq.~\eqref{eq:rel_penalty_n_p2} is more uniform over different similarity scores than Eq.~\eqref{eq:rel_penalty_n_p1} and \eqref{eq:rel_penalty_p2_p1} (compare Fig.~3 in \cite{wang2021understanding}). Since $h(q,p_2) > h(q,n)$ is enforced by the positive ``pull force'' in $\ell_2$, we have $r_{p_2}^{p_1} \gg r_{n}^{p_1}$. Therefore, higher emphasize is put on $h(q,p_1)>h(q,p_2)$ than on $h(q,p_1)>h(q,n)$ in $\ell_1$ and, intuitively, $\ell_2$ emphasizes $h(q,p_2)>h(q,n)$. Thus, $\ell_1$ ensures that $p_1$ and $p_2$ can be discriminated well and $\ell_2$ ensures discrimination between $p_2$ and $n$. In other words, with increasing rank, and thus increasing $\tau$, the focus of the loss gradually shifts from close negatives towards all negatives, effectively increasing with each rank the radius around $q$ at which significant relative penalty results from the negatives. This ``pushing'' force with gradual increasing radius from the respective negatives in combination with the pulling of the respective positives towards $q$ results in ranking.

\begin{figure*}
    \begin{subfigure}{.33\textwidth}
    \centering
    \includegraphics[width=1\linewidth]{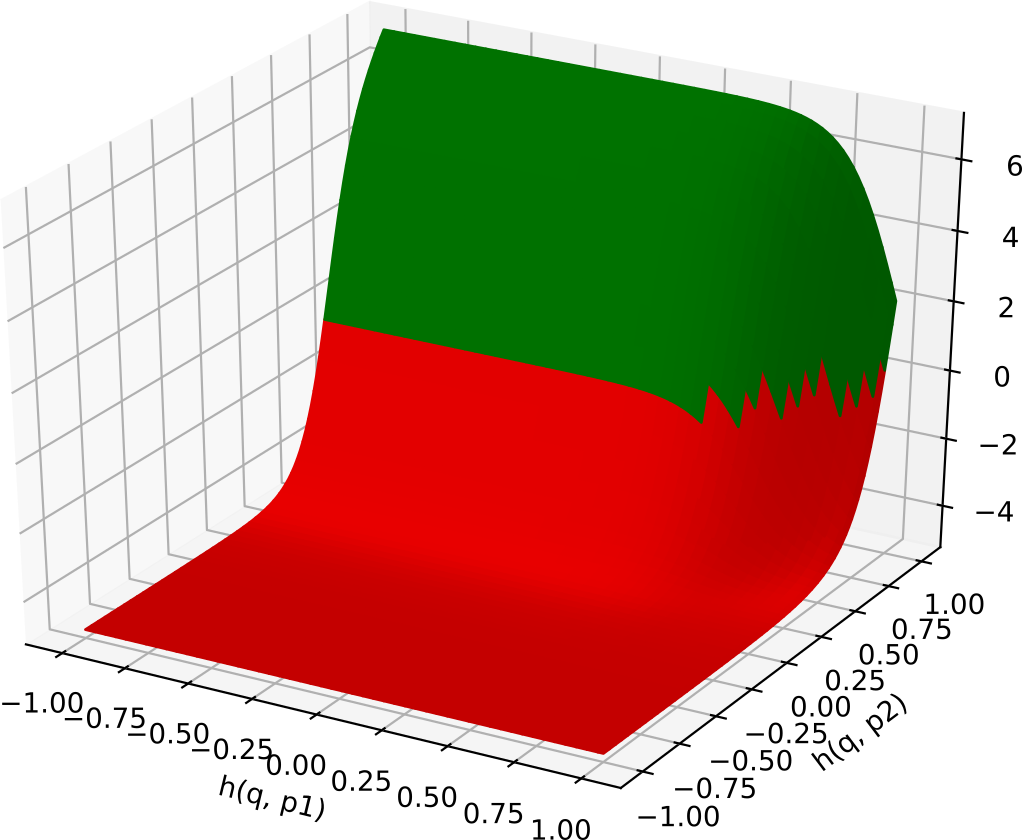}
    \caption{$\tau_1=0.1,\tau_2=0.2$}
    \label{fig:trade-off_t1_01_t2_02}
    \end{subfigure}
    \begin{subfigure}{.33\textwidth}
    \centering
    \includegraphics[width=1\linewidth]{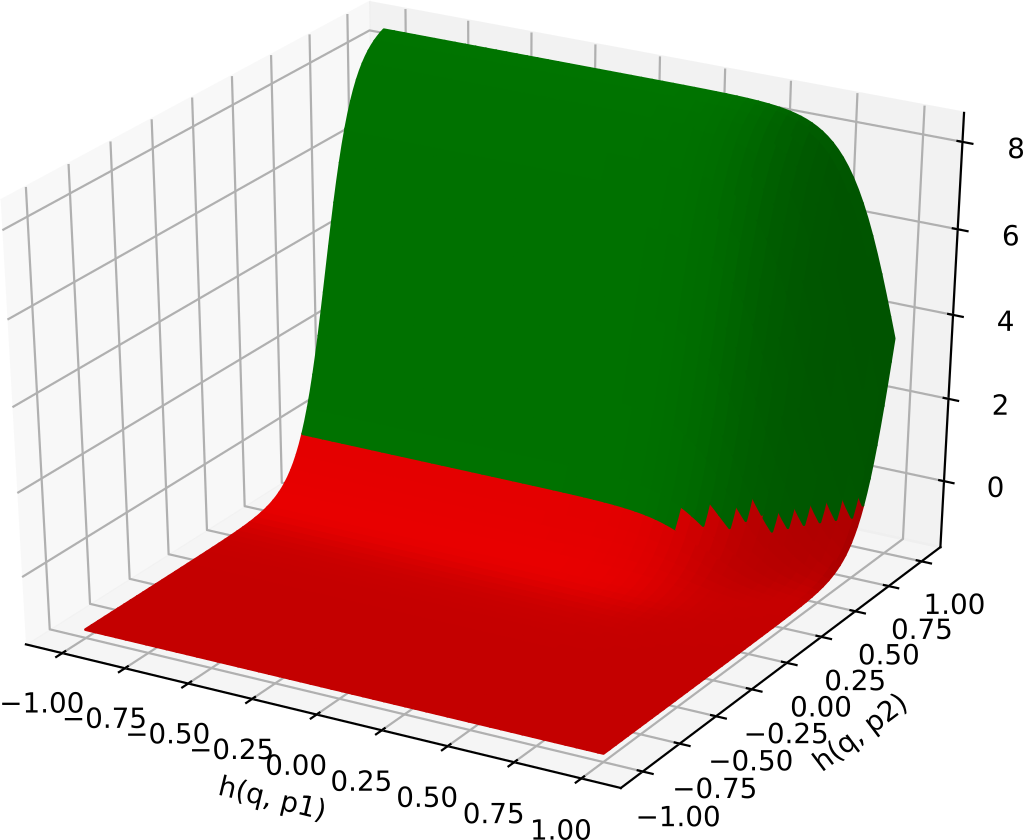}
    \caption{$\tau_1=0.1,\tau_2=0.7$}
    \label{fig:trade-off_t1_01_t2_07}
    \end{subfigure}
    \begin{subfigure}{.33\textwidth}
    \centering
    \includegraphics[width=1\linewidth]{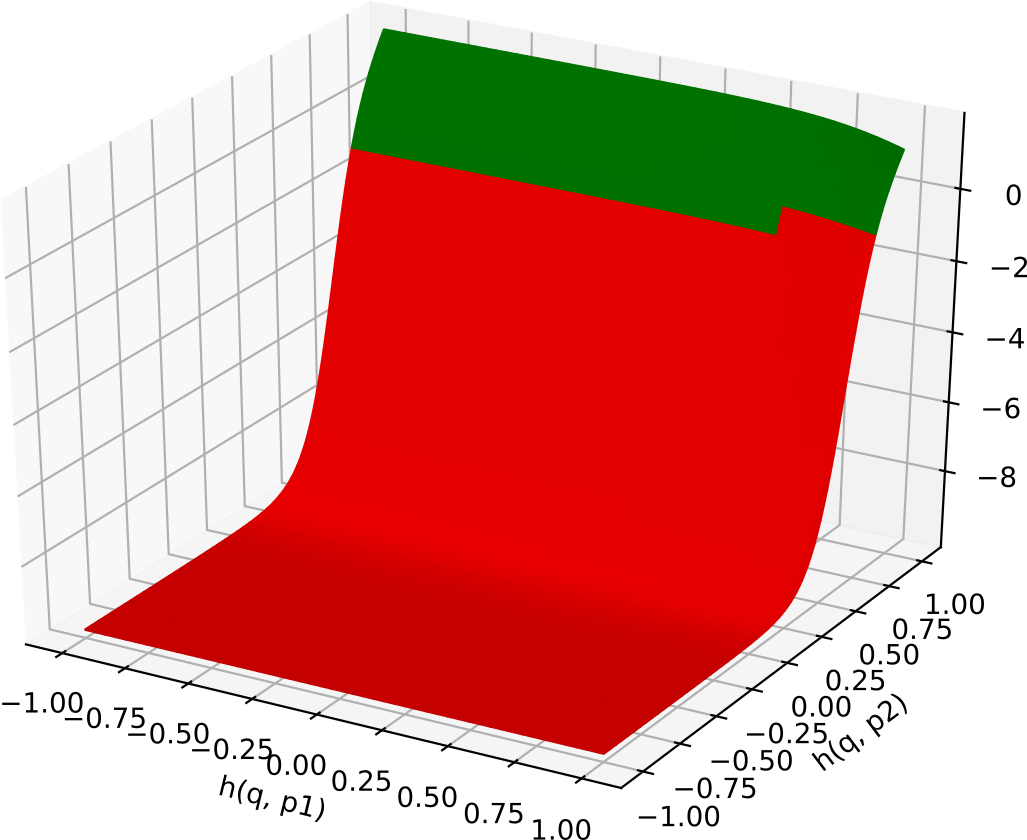}
    \caption{$\tau_1=0.2,\tau_2=0.1$}
    \label{fig:trade-off_t1_02_t2_01}
    \end{subfigure}
    \caption{{\bf Trade-off between opposing RINCE terms for varying $\tau_1$ and $\tau_2$}. We visualize $ \left|\frac{\partial \ell_1}{\partial p_2}\right| - \left|\frac{\partial \ell_2}{\partial p_2}\right|$ for different values of $h(q,p_1)$ and $h(q,p_2)$ ($h(q,n)$ is fixed). We show plots for different $\tau$ values. Red indicates negative values and green positive values. In the red region $h(q,p_2)$ is maximized in the green region minimized. The equilibrium for a fixed $h(q,p_1)$ lies on the boundary of red and green.}
    \label{fig:trade-off}
\end{figure*}

\begin{figure}[t]
    \centering
    \includegraphics[width=\columnwidth]{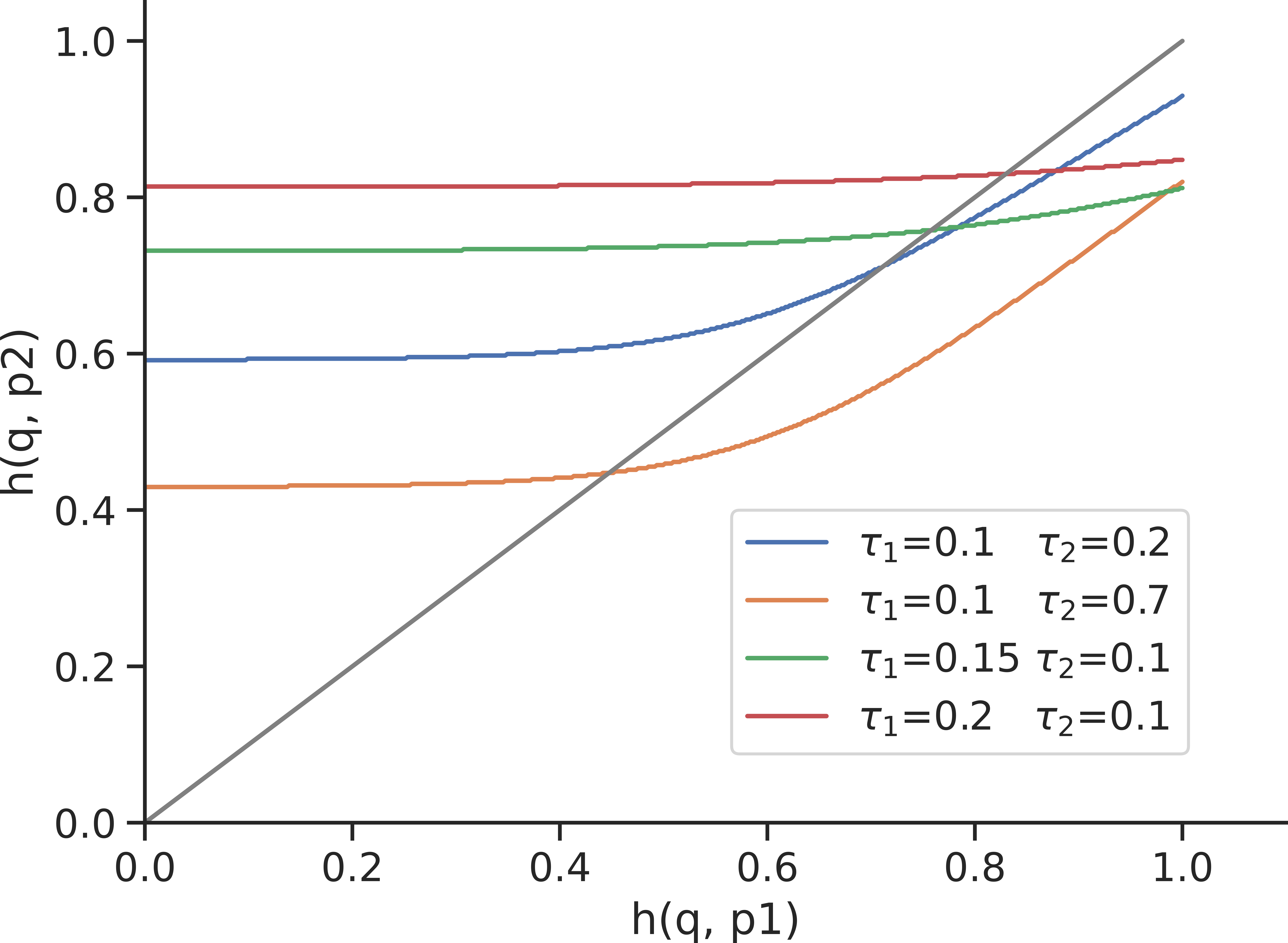}
    \caption{\textbf{Equilibrium lines: Solutions for $K(h(q,p_1), h(q,p_2))=0$ for different values of $\tau_1$ and $\tau_2$.} The lines show all combinations of $h(q,p_1)$ and $h(q,p_2)$ that are solutions of the opposing RINCE terms, i.e.~the gradients are identical except for the sign. The grey line indicates $h(q,p_1)=h(q,p_2)$. For $\tau_1>\tau_2$ it can be seen that with increasing $\tau_1$, the curvature of the equilibrium line steadily decreases resulting in a almost constant value for $h(q,p_2)$ for $\tau_2=0.2$.
    }
    \label{fig:equilibria}
\end{figure}

\begin{table*}[t]
    \centering
    \begin{tabular}{|p{0.95\linewidth}|}
    \hline
    \rule{0pt}{3ex}    
    \textbf{\RINCEsingle.} In the \RINCEsingle loss variant only a single positive is given for each rank, \ie $\mathcal{P}_i=\{p_i\}$. In this case, we use Eq.~\eqref{eq:InfoNCE} for the individual loss terms (Eq.~\eqref{eq:InfoNCE_out} and Eq.~\eqref{eq:InfoNCE_in} are the same for a single positive), and the loss reads:\\
     \vbox{\begin{equation}
         \mathcal{L}_{\text{\RINCEsingle}} = -\sum_{i=1}^{r}\log \frac{\exp(h(q,p_i)/\tau_i)}{\sum_{j=i}^r \exp(h(q,p_j)/\tau_i) + \sum_{n \in \mathcal{N}} \exp(h(q,n)/\tau_i)}.
         \label{eq:rince_single}
         \end{equation}}\\
    \textbf{\RINCEout.}  For the \RINCEout loss variant we take the sum over positives outside of the $\log$ for each rank $i$:\\
     \vbox{\begin{equation}
        \mathcal{L}_{\text{\RINCEout}} = -\sum_{i=1}^{r} \sum_{p \in \mathcal{P}_i}\log \frac{ \exp(h(q,p)/\tau_i)}{\sum_{p \in \{p_i\} \cup \left(\bigcup_{j> i} \mathcal{P}_j\right)} \exp(h(q,p)/\tau_i) + \sum_{n \in \mathcal{N}} \exp(h(q,n)/\tau_i)}.
    \label{eq:rince_out}
    \end{equation}}\\
    \textbf{\RINCEin.} On the other hand, we take the sum over positives inside of the $\log$ for the \RINCEin loss variant:\\
     \vbox{\begin{align}
        \mathcal{L}_{\text{\RINCEin}} = -\sum_{i=1}^{r} \log \frac{\sum_{p \in \mathcal{P}_i} \exp(h(q,p)/\tau_i)}{\sum_{p \in \bigcup_{j\geq i} \mathcal{P}_j} \exp(h(q,p)/\tau_i) + \sum_{n \in \mathcal{N}} \exp(h(q,n)/\tau_i)}.
    \label{eq:rince_in}
    \end{align}} \\
    Note that there is a subtle but noteworthy difference to the $\log$-in version of \cite{khosla2020supervised}, who compute the mean rather than the sum inside the $\log$. As they observe a significantly worse performance of their $\log$-in version, we decide to use the version proposed in \cite{Miech:2020:MILNCE, Han:2020:CoCLR} for all our experiments, including the baseline SCL-in. \\
    \\
    \textbf{\RINCEoutin.} Finally, we consider a combination of the two above: Whenever noise for first rank positives can be expected to be low, while it might be higher for higher rank positives we can use the out-option for the first rank and the in-option for the remaining ranks:\\
     \vbox{\begin{align}
    \begin{split}
        \mathcal{L}_{\text{\RINCEoutin}} 
        = - \sum_{p \in \mathcal{P}_1} \log \frac{ \exp(h(q,p)/\tau_1)}{\sum_{p \in \{p_i\} \cup \left(\bigcup_{j> 1} \mathcal{P}_j\right)} \exp(h(q,p)/\tau_1) + \sum_{n \in \mathcal{N}} \exp(h(q,n)/\tau_1)}\\
        -\sum_{i=2}^{r} \log \frac{\sum_{p \in \mathcal{P}_i} \exp(h(q,p)/\tau_i)}{\sum_{p \in \bigcup_{j\geq i} \mathcal{P}_j} \exp(h(q,p)/\tau_i) + \sum_{n \in \mathcal{N}} \exp(h(q,n)/\tau_i)}.\qquad \quad \,\, 
    \end{split}
    \label{eq:rince_out_in}
    \end{align}
    }\\
    \hline
    \end{tabular}
    \caption{\textbf{Different RINCE loss variants.}
    We assume that an ordered set of positives $\mathcal{P}_1,\dots,\mathcal{P}_r$ is given, where $\mathcal{P}_i$ is the set of positives of rank $i$. We denote the set of negatives by $\mathcal{N}$.}
    \label{tab:all_loss_variants}
\end{table*}

\paragraph{How $\tau$ influences the optimal solution, controls the trade-off between opposing loss terms and why $\tau_i < \tau_{i+1}$ is a good choice for ranking.}

To answer these question we study the trade-off mechanism between negatives in $\ell_1$ and the positives in $\ell_2$, i.e.~the opposing terms in the RINCE loss that lead to the ranking. To investigate the trade-off, we compare the gradient with respect to the negatives $p_2$ in $\ell_1$ with the gradients with respect to the positives in $\ell_2$ (also $p_2$). We want to study which term dominates the entire gradient under which conditions. For this purpose we define 
\begin{equation}
K(h(q,p_1), h(q,p_2)) = \left|\frac{\partial \ell_1}{\partial p_2}\right| - \left|\frac{\partial \ell_2}{\partial p_2}\right|,
\label{eq:gradient_scale}
\end{equation}
which is given by 
\begin{equation}
\begin{split}
&K(h(q,p_1), h(q,p_2))=\\ 
& \frac{1}{\tau_1}\frac{\exp(h(q,p_2)/\tau_1)}{\sum \limits_{n \in \mathcal{N}\cup\mathcal{P}_2}\exp(h(q,n)/\tau_1) + \exp(h(q,p_1)/\tau_1)} \\
    -& \frac{1}{\tau_2}\frac{\sum_{n\in\mathcal{N}}\exp(h(q,n)/\tau_2)}{\sum \limits_{n \in \mathcal{N}}\exp(h(q,n)/\tau_2) + \exp(h(q,p_2)/\tau_2)}
\end{split}
\end{equation}
Intuitively, the value of $K(h(q,p_1),h(q,p_2))$ in Eq.~\eqref{eq:gradient_scale} shows which term dominates the gradient. In our case it even holds that $K(h(q,p_1), h(q,p_2))= \left|\frac{\partial \ell_1}{\partial p_2}\right| - \left|\frac{\partial \ell_2}{\partial p_2}\right| = \frac{\partial \ell_1}{\partial p_2} + \frac{\partial \ell_2}{\partial p_2}$, thus, $K(h(q,p_1), h(q,p_2))$ corresponds to the actual sum of the two gradients, meaning, that $K(h(q,p_1), h(q,p_2))=0$ means the gradients cancel each other out. There exist 3 different cases:
\begin{itemize}
    \item $K>0$: $\left|\frac{\partial \ell_1}{\partial p_2}\right| > \left|\frac{\partial \ell_2}{\partial p_2}\right|$, \ie $\frac{\partial \ell_1}{\partial p_2}$ dominates the gradient and effectively $\ell_1$ minimizes $h(q,p_2)$.
    \item $K<0$: $\left|\frac{\partial \ell_1}{\partial p_2}\right| < \left|\frac{\partial \ell_2}{\partial p_2}\right|$, \ie $\frac{\partial \ell_2}{\partial p_2}$ dominates the gradient and effectively $\ell_2$ maximizes $h(q,p_2)$.
    \item $K=0$: An equilibrium between the opposing terms in $\ell_1$ and $\ell_2$ is found -- neither $\frac{\partial \ell_1}{\partial p_2}$ nor $\frac{\partial \ell_2}{\partial p_2}$ dominates the gradient.
\end{itemize}
We visualize Eq.~\eqref{eq:gradient_scale} in Fig.~\ref{fig:trade-off}. To this end, we model $h(q,n)$ with a Gaussian with $\mu=0.1$ and $\sigma=0.1$ and show different combinations of $\tau$ values, namely $\tau_1=0.1,\tau_2=0.2$ in Fig.~\ref{fig:trade-off_t1_01_t2_02}, $\tau_1=0.1, \tau_2=0.7$ in Fig.~\ref{fig:trade-off_t1_01_t2_07} and $\tau_1=0.2, \tau_2=0.1$ in Fig.~\ref{fig:trade-off_t1_02_t2_01}. For convenience we color negative values in red and positive values in green. The \textit{equilibrium line} separates red and green area. For easier comparison of the equilibrium lines, we visualize them in Fig.~\ref{fig:equilibria}. It shows the value of $h(q,p_2)$ that results in an equilibrium state as a function of $h(q,p_1)$, i.e.~it shows the projection of the equilibrium lines of Fig.~\ref{fig:trade-off} to the $h(q,p_1), h(q,p_2)$ plane. We make the following observations:
\begin{enumerate}
    \item For small values of $h(q,p_1)$ the equilibrium line is close to constant -- the optimization behavior of $h(q,p_2)$ is not influenced by the values of $h(q, p_1)$ -- 
    however, this only occurs very early in training, as $h(q,p_1)$ is maximized without any opposing loss terms.
    \item When $\tau_1<\tau_2$, the equilibrium line indicates that $h(q,p_2)$ grows proportionately to $h(q,p_1)$ (see Fig.~\ref{fig:equilibria} blue and orange).
    \item When $\tau_1<\tau_2$, $h(q,p_1) > h(q,p_2)$ for the interesting regions (see ~grey line in Fig.~\ref{fig:equilibria}). $\tau_2$ can be used to control the minimal similarity of $h(q,p_1)$ required for ranking to appear. Thus, higher ranks require larger $\tau_2$.
    \item When increasing $\tau_2$ for a fixed $\tau_1$, the equilibrium line drops to smaller values overall -- the trade-off is achieved with a smaller similarity of $h(q,p_2)$ (compare Fig.~\ref{fig:equilibria}). Again, this suits higher ranks better.
    \item When $\tau_1>\tau_2$ the curvature of the equilibrium line decreases steadily. For example, with $\tau_1=0.2$ and $\tau_2=0.1$ it is almost constant. The optimization of $h(q,p_2)$ is barely influenced by the values of $h(q,p_1)$ (see green and red line in Fig.~\ref{fig:equilibria}).
\end{enumerate}

Aside from our theoretical justification, we empirically demonstrate that our loss preserves the desired ranking in the feature space, see Fig.~\ref{fig:confusion_mat}.

\subsection{RINCE Loss Variants}
We discussed different RINCE loss variants in the main paper.
Tab.~\ref{tab:all_loss_variants} states the exact equations for the RINCE versions introduced in Tab.~\ref{tab:losses}.

\subsection{Computational Cost}
RINCE only adds a small computational cost to the training pipeline, as only $r-1$ additional computations of the NCE loss function (Eq.~\eqref{eq:InfoNCE}) are necessary. Note, that the dot products of $q$ and all $p \in P$ have to be computed only once and the respective results can be reused for each rank specific loss $\ell_i$ (Eq.~\eqref{eq:ithrank}). In our experiments we did not observe a noticeable difference in overall training time between RINCE and SCL.

\subsection{Hardware details}
Experiments were performed on Nvidia GeForce 1080ti (12GB) and V100 GPUs (32GB), dependent on the respective memory requirements of the models. Contrastive learning for models in the supervised experiment section trained on \cifarhun use a single V100 GPU, while evaluation, \ie training a linear layer, retrieval and OOD experiments use the GeForce 1080ti GPUs. The experiments on ImageNet-100 were run on a V100 GPU. For models in the unsupervised experiments we use a single V100 GPU. The Bosch Group is carbon neutral. Administration, manufacturing and research activities do no longer leave a carbon footprint. This also includes GPU clusters on which the experiments have been performed.

\subsection{Datasets}
\tinyimagenet \cite{le2015tiny} is a small version of ImageNet \cite{deng2009imagenet} comprising only 200 of the 1000 classes with each 500 samples at 64$\times$64 pixel resolution. ImageNet-100 \cite{Tian:2019:CMC} is a subset of ImageNet, consisting of 100 classes of ImageNet at full resolution. We use the RoBERTa \cite{liu2019roberta} model to obtain semantic word embeddings for all class names in ImageNet-100. Second rank positives are based on the word embedding similarity and a predefined threshold. 

\paragraph{Ranking for ImageNet-100.}
\label{par:ranking_ImageNet}
To obtain ranking for  ImageNet-100 we make use of recent progress in natural language processing. We use the RoBERTa \cite{liu2019roberta} implementation provided by \cite{reimers-2019-sentence-bert}\footnote{In particular we use the \emph{stsb-roberta-large} model.} trained for the \emph{semantic textual similarity} benchmark (STSb) \cite{cer2017semeval}. We use this model to embed each class name into a $1024$ dimensional embedding space. Class similarities are computed for each pair using the cosine similarity. A small ablation of the robustness to the similarity score threshold is shown in Tab.~\ref{tab:thr_ablation}. The accuracy is relatively robust on the choice of this threshold. We find $0.45$ to give the best results and use it for the experiments in the main paper.
\begin{table}[h]
    \centering
    \begin{tabular}{lcc}
    \toprule
     Rank2 threshold & Accuracy & AUROC \\
     \midrule
     $0.35$ & $\mathbf{59.78}$ & $60.92$ \\
     $0.40$ & $59.40$ & $61.80$ \\
     $0.45$ & $59.44$ & $\mathbf{62.92}$ \\
     $0.50$ & $59.01$ & $62.05$ \\
     $0.55$ & $58.27$ & $60.54$ \\
     \bottomrule
    \end{tabular}
    \caption{\textbf{Ablation study on the RoBERTa word similarity threshold on \tinyimagenet.} We use different thresholds to define the rank 2 positives. We use \RINCEoutin for pretraining and report accuracy of linear evaluation and AUROC for OOD.}
    \label{tab:thr_ablation}
\end{table}

\subsection{Supervised Contrastive Learning -- Training Details}
Our experiments are based on the Pytorch implementation of \cite{khosla2020supervised}. Common hyper-parameters are used, as given by \cite{khosla2020supervised}. We use ResNet-50 for all models trained on \cifarhun. For ImageNet-100 we use ResNet-18. During contrastive training we use a projection head with a single hidden layer with dimension of $2048$ and an output dimension of $128$.

\paragraph{Optimizer.} 
For both, SCL and RINCE we use stochastic gradient descent with a learning rate of $0.5$, batch size of $512$, momentum of $0.9$, weight decay of $1\mathrm{e}-4$ and a cosine learning rate scheduler. All models are trained for $1000$ epochs. $500$ epochs lead to slightly worse results and results do not change significantly when training for $2000$ instead of $1000$ epochs. Baselines using \textbf{cross-entropy} loss are trained only for $500$ epochs, as we observed that accuracy decreases after epoch $500$. For cross-entropy we use a learning rate of $0.8$, following \cite{khosla2020supervised}. In contrast to \cite{khosla2020supervised} we use a batch size of $512$. We also tested the square root scaling rule for the learning rate, as proposed in \cite{krizhevsky2014one}, but achieve lower accuracy. Our accuracy matches the one reported by \cite{khosla2020supervised}, despite the smaller batch size.

For ImageNet-100 we use a batch size of $768$ and  find a learning rate of $0.3$ to give best results for RINCE and cross-entropy. For SCL we find $0.325$ to give best results. We train all models for $500$ epochs.

\paragraph{Data augmentation.}
We use the same set of standard data augmentations as \cite{khosla2020supervised} in their Pytorch implementation. We create a random crop of size between $20\%$ to $100\%$ of the initial image with random aspect ration between $3/4$ and $4/3$ of the initial aspect ratio. The resulting crop is scaled to $32\times32$ pixels for \cifarhun and $224\times224$ for ImageNet-100. We flip images with a probability of $0.5$ and apply color jitter randomly selected from $[0.6,1.4]$ for brightness, contrast and saturation and apply jitter to the hue from $[-0.1, 0.1]$ with a probability of $0.8$. Finally, we convert the image to grayscale with probability of $0.2$. Cross-entropy does not use color jitter and random grayscaling, but except from that uses the same augmentations. Cross-entropy strong augmentation (cross-entropy s.a.) uses the exact same augmentations as RINCE and SCL.

\paragraph{Memory bank and MoCo.}
\begin{table}[t]
    \centering
    \begin{tabular}{lc}
    \toprule
     memorybank size & Accuracy\\
     \midrule
     2048 & 77.03 \\
     4096 & \textbf{77.46} \\
     8092 & 77.25 \\
     \bottomrule
    \end{tabular}
    \caption{\textbf{Ablation study on the size of the memorybank on \cifarhun.} We use \RINCEoutin for pretraining and train a linear layer on top of the frozen network with cross-entropy. }
    \label{tab:membank_ablation}
\end{table}
To obtain a memory efficient solution, that can run on a single GPU we use a memory bank with the MoCo trick \cite{He:2019:MoCo}. Our search space for the memory bank size is inspired by the memory bank size used in an ablation study of \cite{khosla2020supervised}. Since we are training on \cifarhun with only $100$ classes in comparison for $1000$ for ImageNet, we also try a smaller value. In initial experiments we compared three memory bank sizes: $2048$, $4096$ and $8192$, see Tab.~\ref{tab:membank_ablation}. Differences were minor, but slightly better for $4096$. We use this value for all models trained on \cifarhun. For ImageNet-100 we use 8192, without further ablating it.

In MoCo training a second encoder network is used to get the representation for positives and negatives. The weights of this so-called key-encoder are a momentum-based moving average of the query encoder weights. We choose a value of $0.99$ as momentum, without further ablating it.

\paragraph{Temperature $\tau$.}

\begin{table}[h]
    \centering
    \begin{tabular}{lc}
    \toprule
     Rank2 temperature & Accuracy\\
     \midrule
     $\tau_2$=0.125 & 75.98 \\
     $\tau_2$=0.175& 76.44\\
     $\tau_2$=0.225& \textbf{77.18}\\
     $\tau_2$=0.25& 76.87 \\
     \bottomrule
    \end{tabular}
    \caption{\textbf{Ablation study on $\tau_2$ on \cifarhun.} We use \RINCEin for pretraining and train a linear layer on top of the frozen network with cross-entropy. }
    \label{tab:tau_max_ablation}
\end{table}
A critical parameter for the \infoNCE loss is the temperature $\tau$. RINCE requires determining a range of $\tau$ values. In practice, we found that starting with common values for $\tau_1$ and then linearly spacing $\tau_i$, $i>1$ works well. Effectively, this doubles the search effort. For SCL and RINCE we use the identical $\tau$ for rank 1. We tested $0.07$ and $0.1$ and found $0.1$ to work slightly better. For RINCE, a $\tau$ needs to be selected for each rank. For $\tau_2$ we searched for a good value over the range $[0.125, 0.25]$ for \RINCEin, while keeping $\tau_1$ fixed to $0.1$. We found $\tau_2=0.225$ to work best. The corresponding ablation study on the sensitivity to the $\tau_2$ is shown in Tab.~\ref{tab:tau_max_ablation}.

\paragraph{Randomness.}

Whenever we provide mean and standard deviation we set the following random seeds: $123$, $546$ and $937$. We set them for Numpy and Pytorch individually.

\begin{table*}[h!]
    \centering
    \begin{tabular}{l|ccc|cc}
    \toprule
          & \multicolumn{3}{c|}{Classification \cifarhun} & \multicolumn{2}{c}{OOD} \\
         Smoothing factor& Accuracy & R@1 fine & R@1 superclass &  $\mathcal{D}_{out}$: \cifarten & $\mathcal{D}_{out}$: \tinyimagenet \\
         \midrule
         $\alpha=0.1$ & \textbf{75.66 $\pm$ 0.29} & \textbf{75.39 $\pm$ 0.33} & 85.42 $\pm$ 0.14 & 73.85 $\pm$ 0.19 & 80.04 $\pm$ 0.78  \\
         $\alpha=0.2$ &\textbf{75.66 $\pm$ 0.10} & 75.04 $\pm$ 0.06 & 85.38 $\pm$ 0.20 & 74.20 $\pm$ 0.23 & 79.84 $\pm$ 0.04 \\ 
         $\alpha=0.3$ & \textbf{75.66 $\pm$ 0.27} & 74.90 $\pm$ 0.06 &\textbf{85.59 $\pm$ 0.12} & \textbf{74.35 $\pm$ 0.65} & \textbf{80.10 $\pm$ 0.77} \\
    \bottomrule
    \end{tabular}
    \caption{\textbf{Label Smoothing, effect of} $\mathbf{\alpha}$. Hyper-parameter search for the smoothing factor $\alpha$, which determines how strongly the one-hot-vector is smoothed. We report the mean and standard deviation over three runs.
    }
    \label{tab:smoooth_labels}
\end{table*}

\begin{table*}[h!]
    \centering
    \begin{tabular}{l|ccc|cc}
    \toprule
          & \multicolumn{3}{c|}{Classification \cifarhun} & \multicolumn{2}{c}{OOD} \\
         loss weight& Accuracy & R@1 fine & R@1 superclass &  $\mathcal{D}_{out}$: \cifarten & $\mathcal{D}_{out}$: \tinyimagenet \\
         \midrule
        $\lambda=0.1$ & 73.85 $\pm$ 0.54 & 73.15 $\pm$ 0.80 & 81.37 $\pm$ 0.90 & \textbf{78.15 $\pm$ 0.49} & 77.91 $\pm$ 0.48 \\
        $\lambda=0.2$ & \textbf{74.08 $\pm$ 0.40} & 73.62 $\pm$ 0.31 & 81.92 $\pm$ 0.21 & 77.99 $\pm$ 0.07 & 78.35 $\pm$ 0.39 \\
        $\lambda=0.3$ & 74.05 $\pm$ 0.48 &\textbf{ 73.67 $\pm$ 0.29} &\textbf{ 82.39 $\pm$ 0.22} & 78.13 $\pm$ 0.11 & \textbf{78.97 $\pm$ 0.21} \\

    \bottomrule
    \end{tabular}
    \caption{\textbf{Two heads: Effect of loss weight $\mathbf{\lambda}$}. Hyper-parameter search for the loss weight $\lambda$, which controls how much the superclass classification contributes to the loss. We report the mean and standard deviation over 3 runs.}
    \label{tab:two_heads}
\end{table*}

\paragraph{Training a linear layer.}

After contrastive training we remove the MLP projection head and replace it with a single linear layer. We freeze the entire network, including the batch norm parameters and only train the weights of the linear layer with cross-entropy. For linear evaluation we use stochastic gradient descent with a learning rate of $5$ and a batch size of $512$ for \cifarhun and for ImageNet-100. We decay the learning rate at epoch $60$, $75$ and $90$ with a decay rate of $0.2$.

\subsection{Baselines -- Supervised Contrastive Learning}

Additionally to the SCL and cross entropy baselines, we provide 6 additional supervised baselines: \textit{label smoothing}, \textit{two heads}, \textit{SCL two heads}, \textit{Triplet}, \textit{Hierarchical triplet} and \textit{Fast AP}. The first two baselines are trained with cross-entropy. The hyper-parameters used are identical to those of the cross-entropy baseline. For SCL two heads we pick the same hyper-parameters as for SCL. Hyper-parameters that are new for the respective methods are determined with a parameter search. We always pick the model that results in highest accuracy after training a linear probe on top of the frozen features.

\paragraph{Label smoothing.}
Label smoothing simply converts the one-hot-vectors used as target in the cross-entropy loss into a probability distribution over the target labels by assigning some of the probability to the other labels. In contrast to basic label smoothing and to use the same information as provided to RINCE, \ie which classes belong to the same superclass, we do not distribute the probability mass to all classes, but only to those within the same superclass. Within the superclass we distribute the probability mass uniformly. A critical parameter is the smoothing factor $\alpha$, which denotes the fraction of the probability mass removed from the target class and distributed among the remaining classes. We tested three values and picked the one resulting in best accuracy on \cifarhun. All models are shown in Tab.~\ref{tab:smoooth_labels}.

\paragraph{Two heads.}
The second simple baseline we compare to is referred to as \emph{two heads}. To make use of the information provided by the superclasses we add a second classification head to the ResNet-50 backbone. While the first head is trained as before for fine label classification on \cifarhun, the second head is trained to predict the superclass labels. The loss is given by $\mathcal{L}= (1-\lambda) \mathcal{L}_{\text{fine}} + \lambda \mathcal{L}_{\text{superclass}}$. Critical is the weighting parameter $\lambda$. To find a good value we ran a small hyper-parameter search. The results are shown in Tab.~\ref{tab:two_heads}. Again, we picked the model with highest classification accuracy on \cifarhun fine labels for the main paper. Note, that $\lambda=0.1$ leads to best OOD scores on $\mathcal{D}_{out}$: \cifarten, however, results on classification and retrieval are below vanilla cross-entropy (compare Tab.~\ref{tab:both}).

\paragraph{SCL two heads.}
\begin{table*}[h]
    \centering
    \begin{tabular}{ll|ccc|cc}
    \toprule
          & & \multicolumn{3}{c|}{Classification \cifarhun} & \multicolumn{2}{c}{OOD} \\
          & loss weight& Accuracy & R@1 fine & R@1 superclass &  $\mathcal{D}_{out}$: \cifarten & $\mathcal{D}_{out}$: \tinyimagenet \\
         \midrule
        \multirow{4}{*}{SCL-in} & $\lambda=0.1$ & 75.92  & 72.76 & 81.67 &74.93 & 77.95\\
        & $\lambda=0.2$ &  76.91 & \textbf{76.91} & 74,40 & 75.26 & \textbf{79.55}\\
        & $\lambda=0.3$ &  \textbf{76.97} & 74.46 & 83,39 & 75.36 & 79.11\\
        & $\lambda=0.4$ & 76.44 & 75.07 & \textbf{84.02} & \textbf{75.83} & 79.52\\
        \midrule
        \multirow{5}{*}{SCL-out} & $\lambda=0.1$ &  75.94 & 72.76 & 81.67 & 74.93 & 77.95\\
        & $\lambda=0.2$ &  75.82 & 73.68 & 82.25 & 75.26 & 79.74\\
        & $\lambda=0.3$ &  76.47 & 74.36 & 83.27 & 74.52 & 79.00\\
        & $\lambda=0.4$ &  \textbf{76.89} & 74.37 & 83.60 & 74.98 & 79.33\\
        & $\lambda=0.5$ &  76.32 & \textbf{75.09} & \textbf{84.02} & \textbf{75.76} & \textbf{79.86}\\
    \bottomrule
    \end{tabular}
    \caption{{\textbf{SCL two heads: Effect of loss weight $\mathbf{\lambda}$}. Hyper-parameter search for the loss weight $\lambda$, which controls how much the superclass classification contributes to the loss.}}
    \label{tab:scl_two_heads}
\end{table*}
Similarly, as for \emph{two heads}, for \emph{SCL two heads} we train simultaneously on fine and coarse labels. Instead of using cross-entropy we use the SCL loss. We follow the normla \emph{SCL} setting with a projection head. In contrast to \emph{SCL} we add a second projection head. The first head is trained like SCL on the fine labels, while the second is trained on the superclass labels. Similar to \emph{two heads}, the loss is given by $\mathcal{L}= (1-\lambda) \mathcal{L}_{\text{fine}} + \lambda \mathcal{L}_{\text{superclass}}$. Ablation on hyper-parameter $\lambda$ can be found in Tab.~\ref{tab:scl_two_heads}.

\paragraph{Triplet Loss.}
We use the triplet margin loss given by 
\begin{equation}
\mathcal{L}_\text{triplet} = \text{max}\{ d(q,p) - d(q,n) + m, 0\},
\end{equation} 
where $d(\cdot, \cdot)$ is the euclidean distance and $m$ the margin. We adapt it to the ranking setting by choosing different margins based on the rank, We find $m_1=0.5$ and $m_2=1$ to work best in our setting. Further, we find a learning rate of $0.75$ to yield best results.

\paragraph{Hierarchical Triplet.}
Hierarchical Triplet \cite{ge2018deep} is a method for supervised learning. With the goal of hard example mining, the method learns a class hierarchy and draws samples from similar classes more frequently than dissimilar ones. Besides that, Hierarchical Triplet defines an adaptive violate margin, which depends on the class hierarchy. For a fair comparison to our setting, we do not learn the hierarchy. Instead, we use the ground truth hierarchy given by the \cifarhun superclasses, similar as for RINCE. We use a batch size of $512$.

Hierarchical triplet performs hard example mining by sampling related classes with higher probability. Hard example mining is controlled with the following parameters: $l'$ denotes the number of random classes per batch, $c$ denotes the number of samples drawn from the closest class for each sample and $t$ denotes the number of randomly drawn samples. Thus, $\text{batch\_size}=l' + l'c+t$. We find $l'=30$, $c=10$ and $t=182$ to work well.

\paragraph{Fast AP.}
Fast AP \cite{cakir2019deep} is a metric learning method tailored towards learning to rank. The loss directly optimizes Average Precision (AP) and is shown to result in high retrieval scores. The method uses differential histogram binning to efficiently approximate AP. Besides that, it introduces a special batch sampling strategy, which first samples categories (Cifar superclasses) and then for each category a number of samples (Cifar fine labels). We stick to the sampling strategy as proposed in the paper and sample per batch 2 categories. We ran a hyper-parameter search on the number of histogram bins and the learning rate. We found 5 histograms and a learning rate of 0.1 to work best in our setting. The remaining training details are identical to RINCE and the other baselines reported in this paper.

\begin{figure*}[h!]
    \centering
    \includegraphics[width=0.85\textwidth]{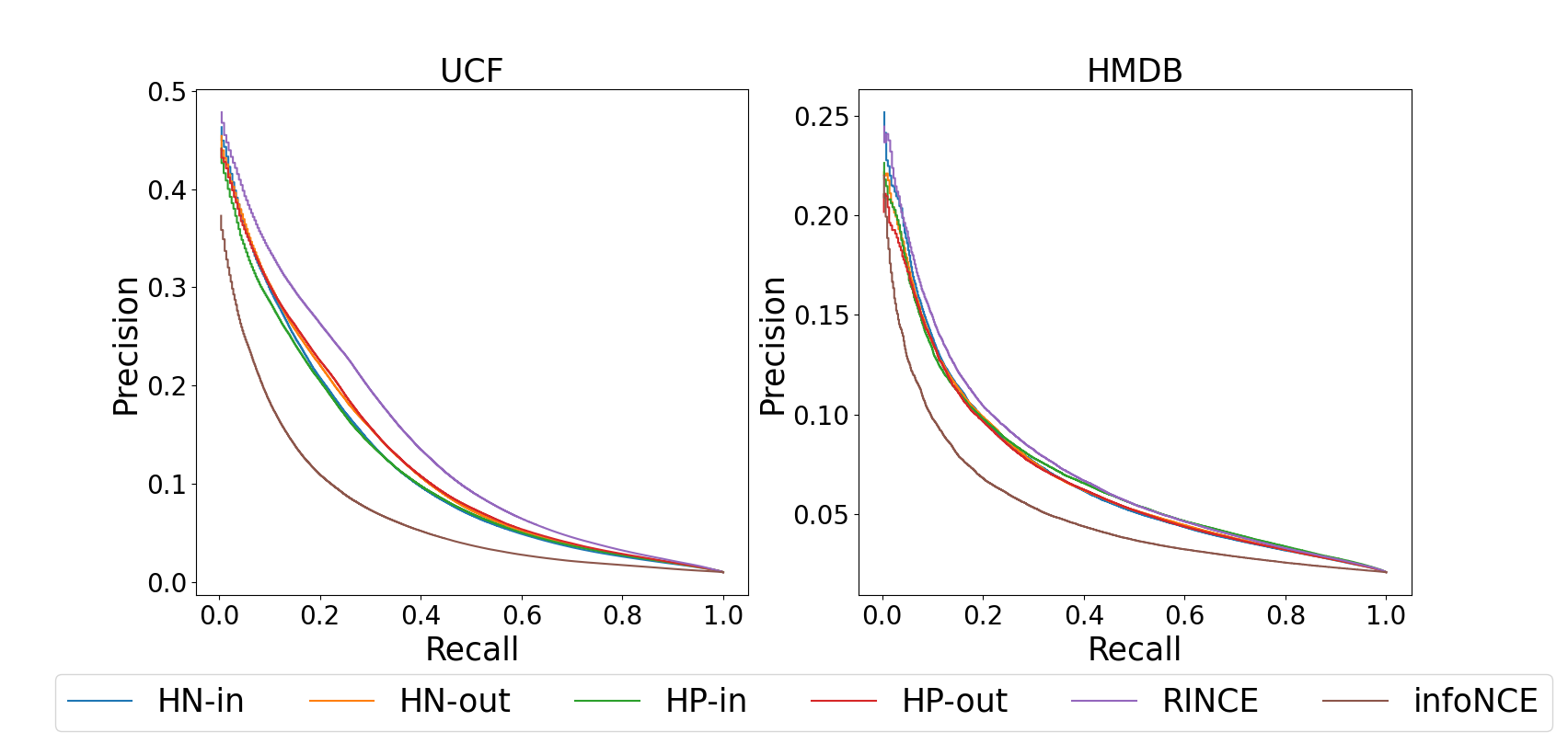}
    \caption{{\bf Precision-Recall Curves on UCF and HMDB.} We show the precision-recall curves for the \infoNCE baseline, hard negatives (HN), hard positives (HP) and RINCE on the two action recognition datasets UCF and HMDB. We observe that both the in- and out-option of HN and HP improve over \infoNCE, while RINCE performs best.}
    \label{fig:PR_temporalranking}
\end{figure*}

\begin{figure*}[t]
    \centering
    \begin{subfigure}{.235\textwidth}
  \centering
\includegraphics[width=\textwidth]{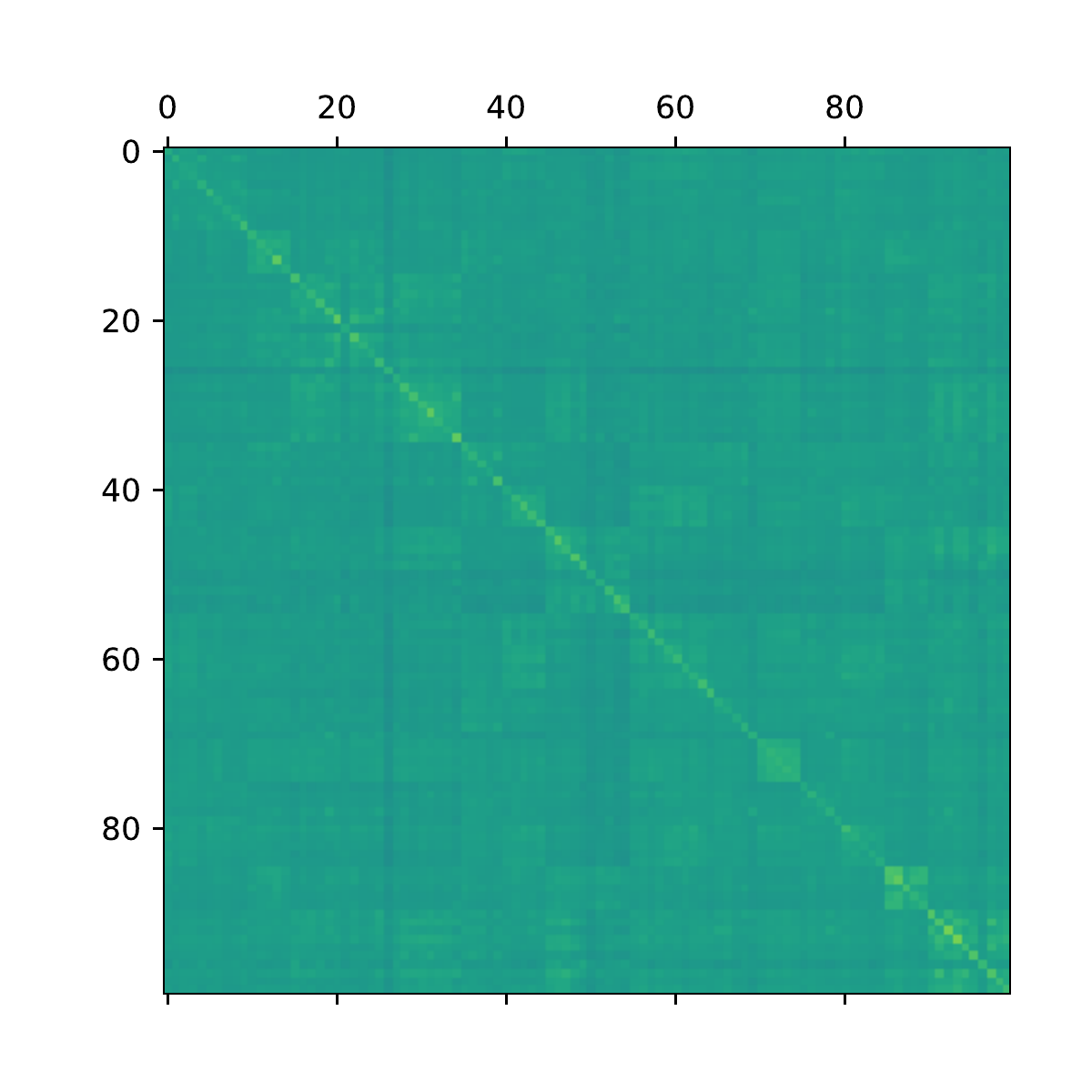}
  \caption{SCL-in before MLP}
  \label{fig:conf_mat_scl_in_before}
    \end{subfigure}%
    \begin{subfigure}{.235\textwidth}
  \centering
  \includegraphics[width=\textwidth]{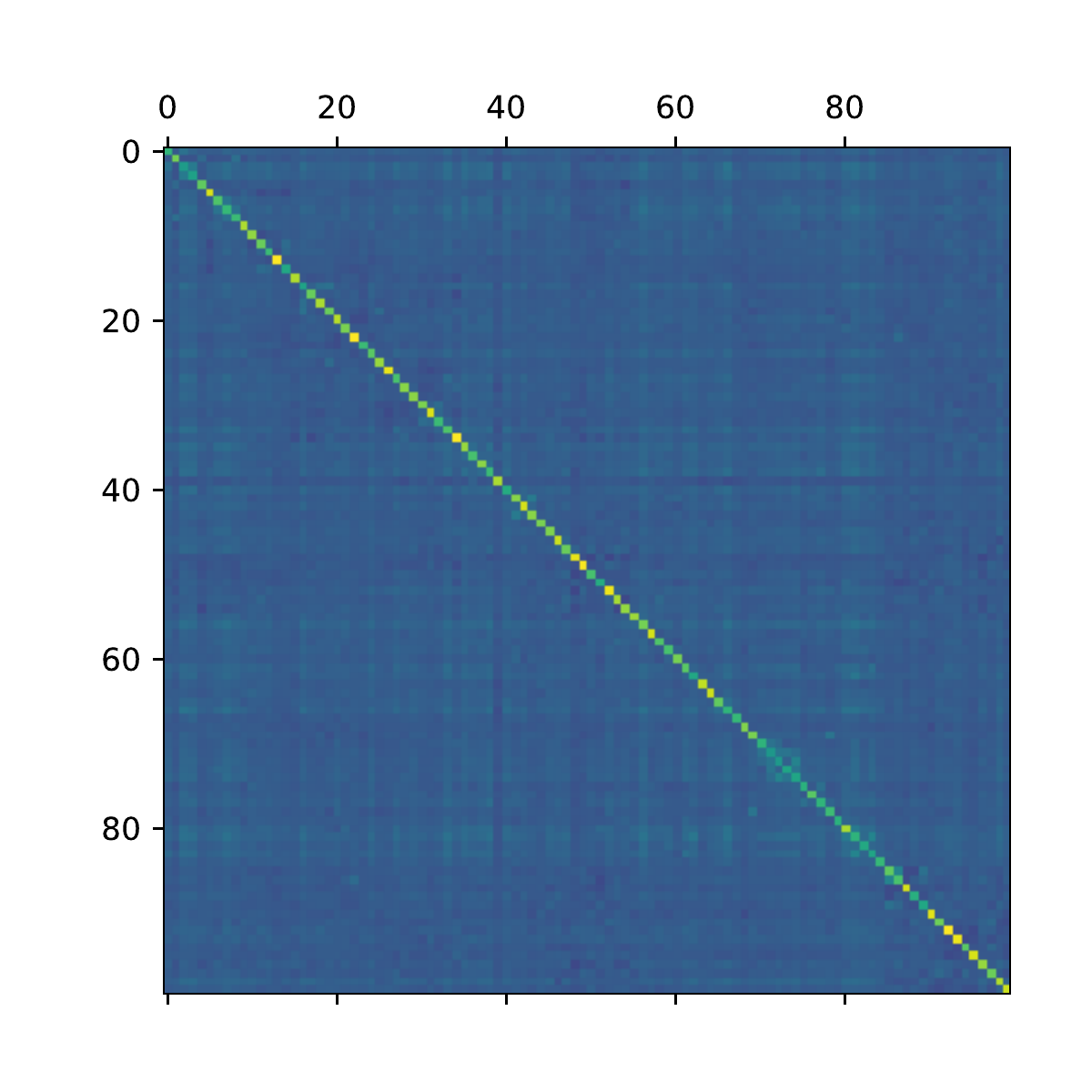}
  \caption{SCL-in after MLP}
  \label{fig:conf_mat_scl_in_after}
\end{subfigure}
\begin{subfigure}{.235\textwidth}
    \centering
    \includegraphics[width=\textwidth]{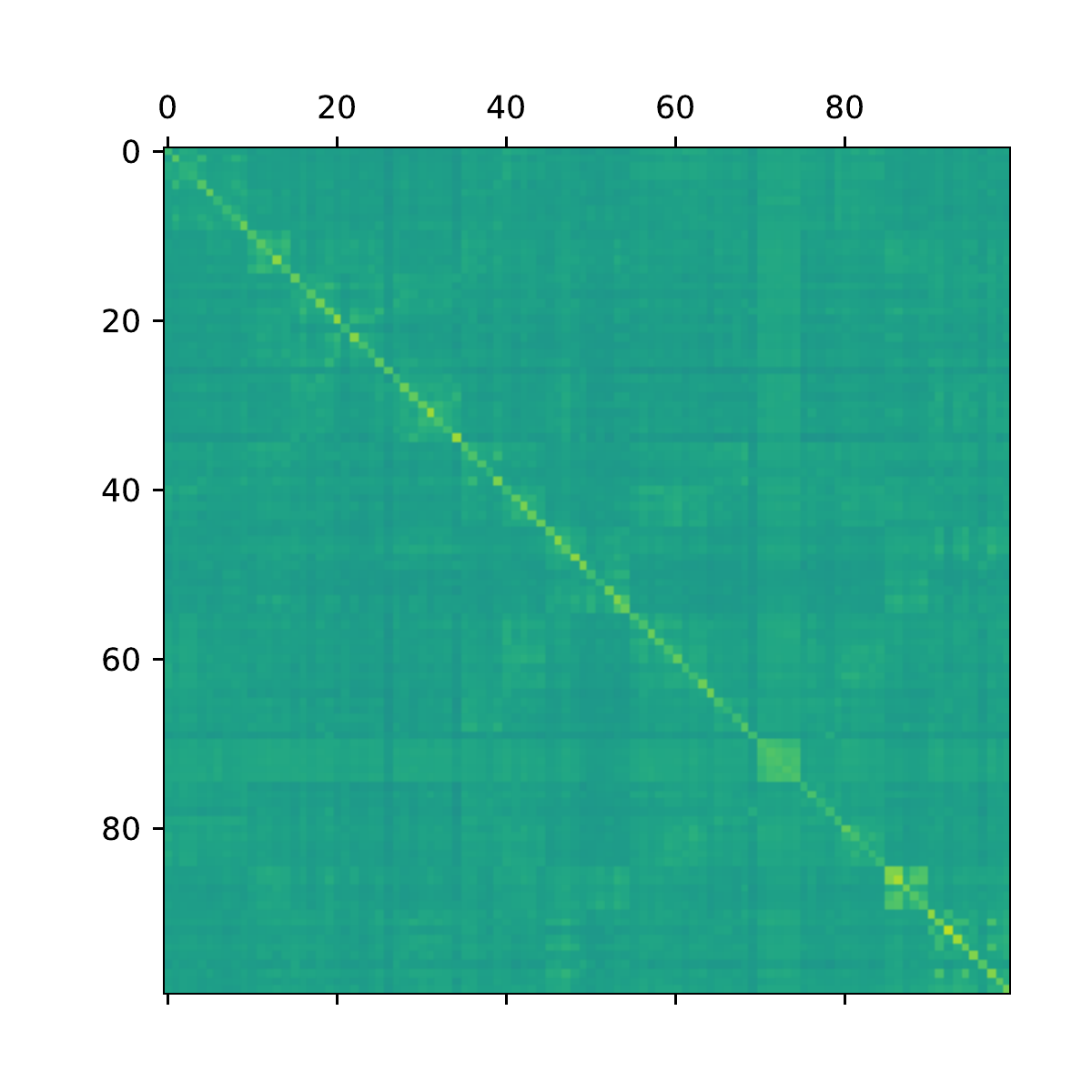}
    \caption{\RINCEin before MLP}
    \label{fig:conf_mat_rince_in_before}
    \end{subfigure}
\begin{subfigure}{.235\textwidth}
    \centering
    \includegraphics[width=\textwidth]{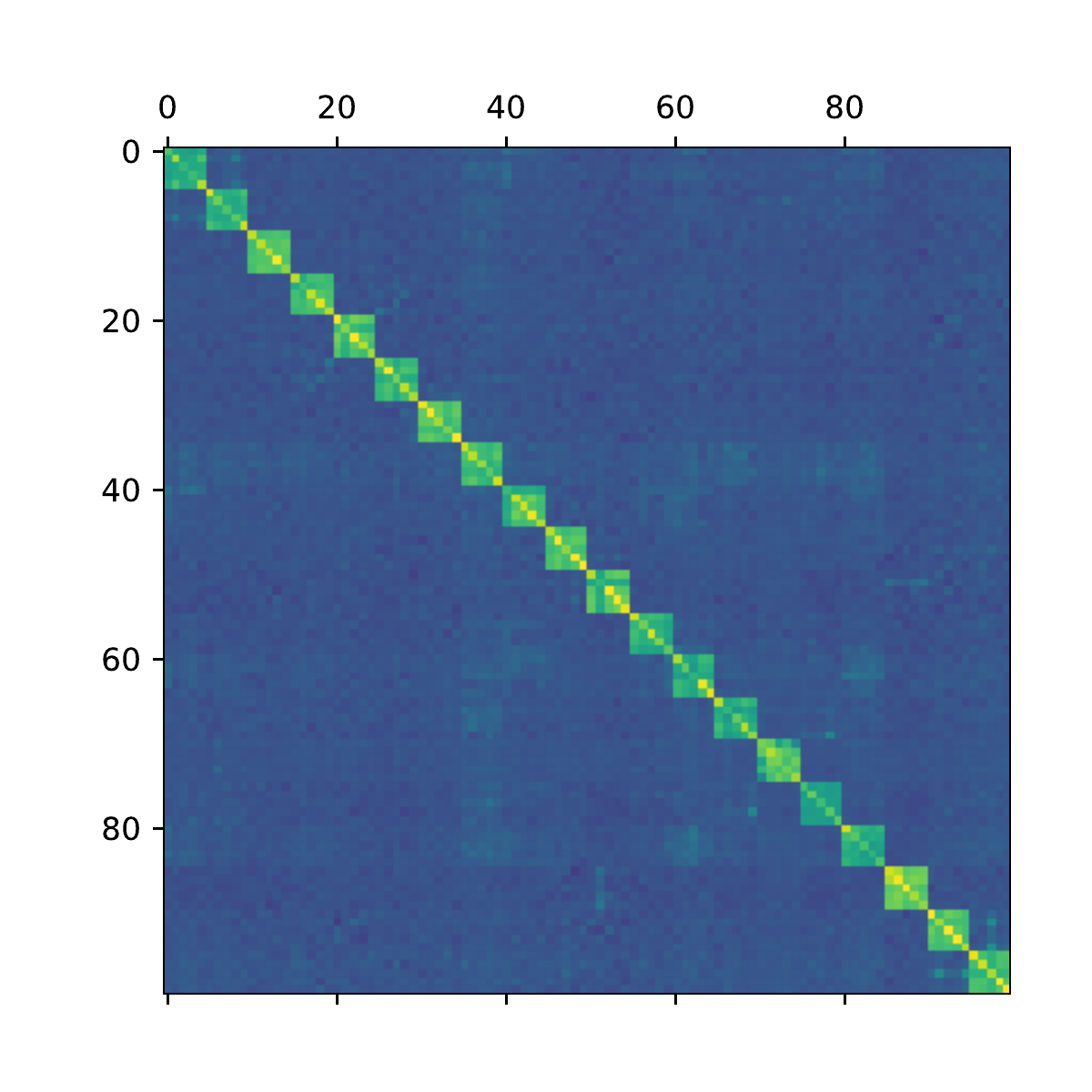}
    \caption{\RINCEin after MLP}
    \label{fig:conf_mat_rince_in_after}
    \end{subfigure}
\begin{subfigure}{.04\textwidth}
    \centering
    \includegraphics[width=1\linewidth]{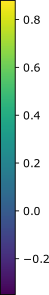}
    \label{fig:conf_mat_colorbar}
    \end{subfigure}    
\caption{\textbf{Similarity matrix on \cifarhun classes before and after the MLP head.} Classes are sorted, such that superclasses are grouped together. Similarity values are the average cosine similarity between classes. Similarity is larger for RINCE after the MLP, therefore ranking is learned by the MLP.}
\label{fig:confusion_mat}
\end{figure*}

\subsection{Out-of-Distribution Detection}
After training we follow the setup of \cite{winkens2020contrastive} and fit $C$ $n$-dimensional class-conditional multivariate Gaussian to the embedded training samples with \cite{scikit-learn}, where $n$ is the dimension of the embedding space and $C$ denotes the number of classes in $\mathcal{D_\text{in}}$.
The OOD score is defined as 
\begin{equation}
    s(x) = \max_c(\log(\textit{L}_c(x))),
\end{equation}
where $\textit{L}_c$ denotes the likelihood function of the Gaussian for class $c$.
Note, that this approach does not require any data labelled as OOD sample and can be applied out-of-the-box.
We use this approach for all our baselines, \ie cross-entropy, label smoothing, two-heads, SCL-in superclass, SCL-in and SCL-out.

For evaluation, we compute the area under the receiver operating characteristic curve (AUROC). Note that this metric is independent of any threshold and can be directly used on the OOD-scores. An intuitive interpretation of this metric is as the probability that a randomly picked in-distribution sample gets a higher \emph{in-distribution score} than an OOD sample.

\subsection{Unsupervised RINCE -- Video Experiment}

\paragraph{Implementation details.}
We use a 3D-Resnet18 backbone \cite{Hara:2018:3DResnet} in all experiments and pool the feature map into a single $512$-dimensional feature vector. The MLP head $g$ has $512$ hidden units with ReLu activation. Note that the MLP head is removed after self-supervised training and will not be transferred to downstream tasks. We use the Adam optimizer \cite{Kingma:2015:Adam} with weight decay $1\mathrm{e}{-5}$, a batch size of $128$ and an initial learning rate of $1\mathrm{e}{-3}$, that is decreased by a factor of $10$ when the validation loss plateaus, and train for $200$ epochs. We use a memory bank size of $65.536$~\cite{He:2019:MoCo} and do the momentum update with $m=0.99$. We use a temperature parameter $\tau=0.1$ for the baselines (\infoNCE, hard positives, hard negatives) and $\tau_1=0.1, \tau_2=0.15, \tau_3=0.2$ for RINCE.

\paragraph{Finetuning.}
Downstream performances are reported on split 1 of UCF and HMDB. We use pretrained weights of the baselines and RINCE to initialize a 3D-Resnet18, add a randomly initialized linear layer and dropout with a dropout rate of $0.9$, and finetune everything end-to-end using cross entropy. We finetune the models for $500$ epochs using the Adam optimizer with weight decay $1\mathrm{e}{-5}$ and a learning rate of $1\mathrm{e}{-4}$ that is reduced by a factor of $10$ when the validation loss plateaus.

\paragraph{Frame-, Shot- and Video-level Positives.}
We sample short clips each consisting of $16$ frames sampled with a temporal stride of $3$ for $x_f$, $x_s$ and $x_v$. We ensure a gap of at least $48$ frames between $x_v$ and the other two positives $x_f$ and $x_s$. We augment each clip with a set of standard video augmentations: random sized crop of size $128\times 128$, horizontal flip, color jittering and random color drop.

\paragraph{Precision-Recall Curves on UCF and HMDB.}
We do the same retrieval evaluation as previously described for videos of UCF and HMDB and provide the resulting precision-recall curves in Fig.~\ref{fig:PR_temporalranking}. We observe that the hard positive (HP) and hard negative (HN) baselines improve over \infoNCE, and RINCE outperforms all baselines. This is in line with our findings in Tab.~\ref{table:temporal_ranking}.

\subsection{Does RINCE rank samples?} 
\begin{figure*}
    \centering
    \begin{subfigure}{.49\textwidth}
  \centering
  \includegraphics[width=1\linewidth]{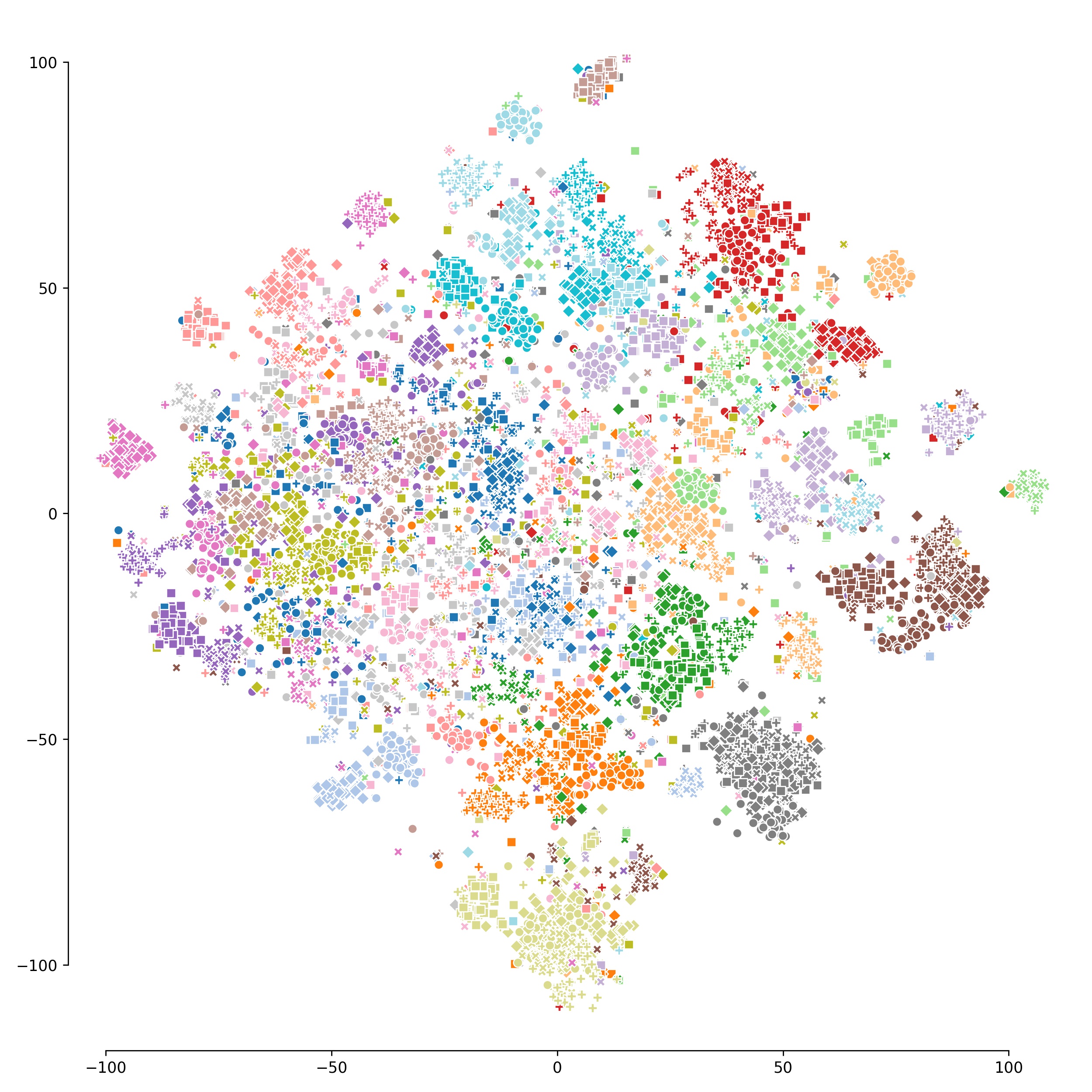}
  \caption{SCL-in}
  \label{fig:tsne2_scl}
  \hfill
    \end{subfigure}
\begin{subfigure}{.49\textwidth}
  \centering
  \includegraphics[width=1\linewidth]{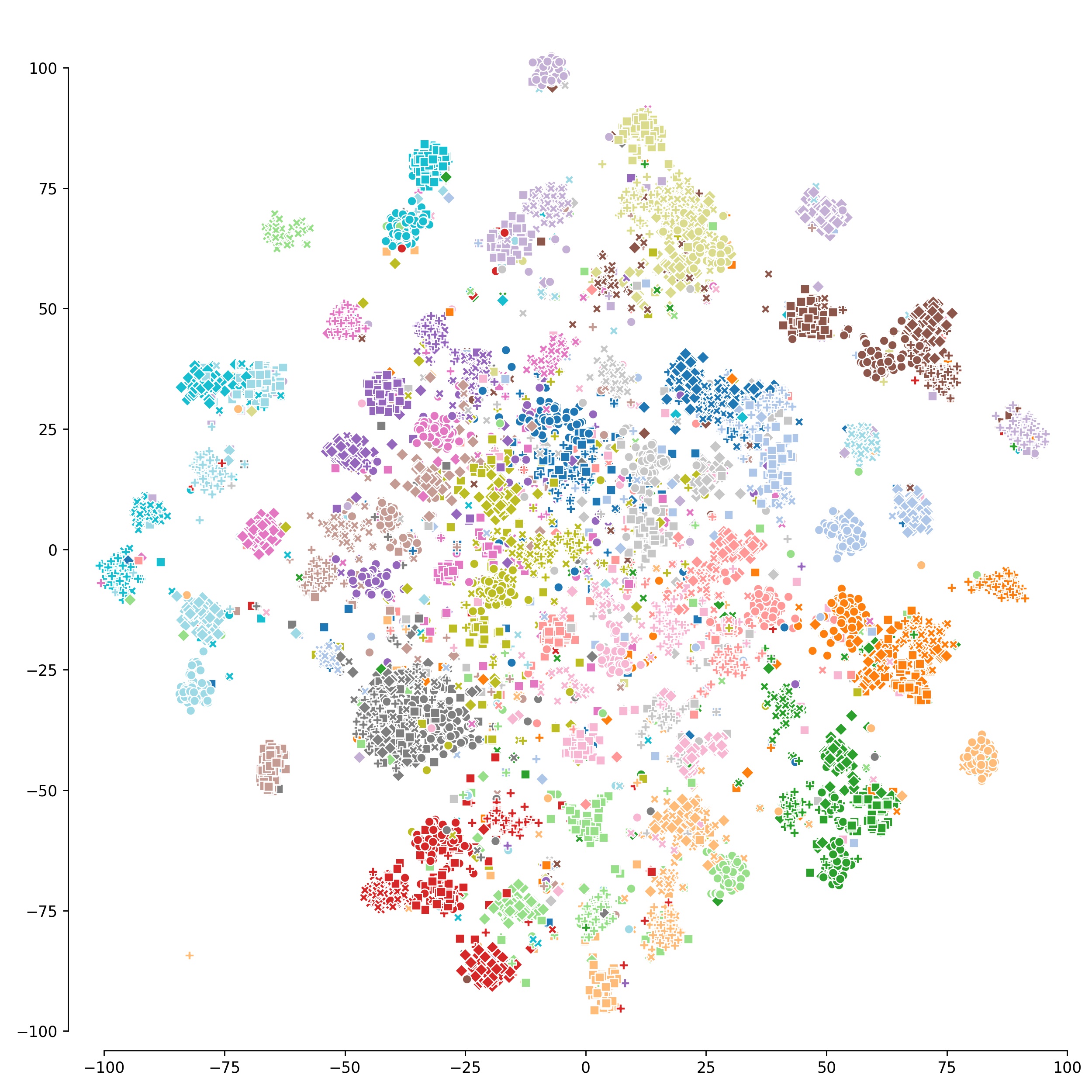}
  \caption{\RINCEoutin}
  \label{fig:tsne2_RINCE}
\end{subfigure}

\centering
\begin{subfigure}{0.49\textwidth}
  \centering
  \includegraphics[width=1\linewidth]{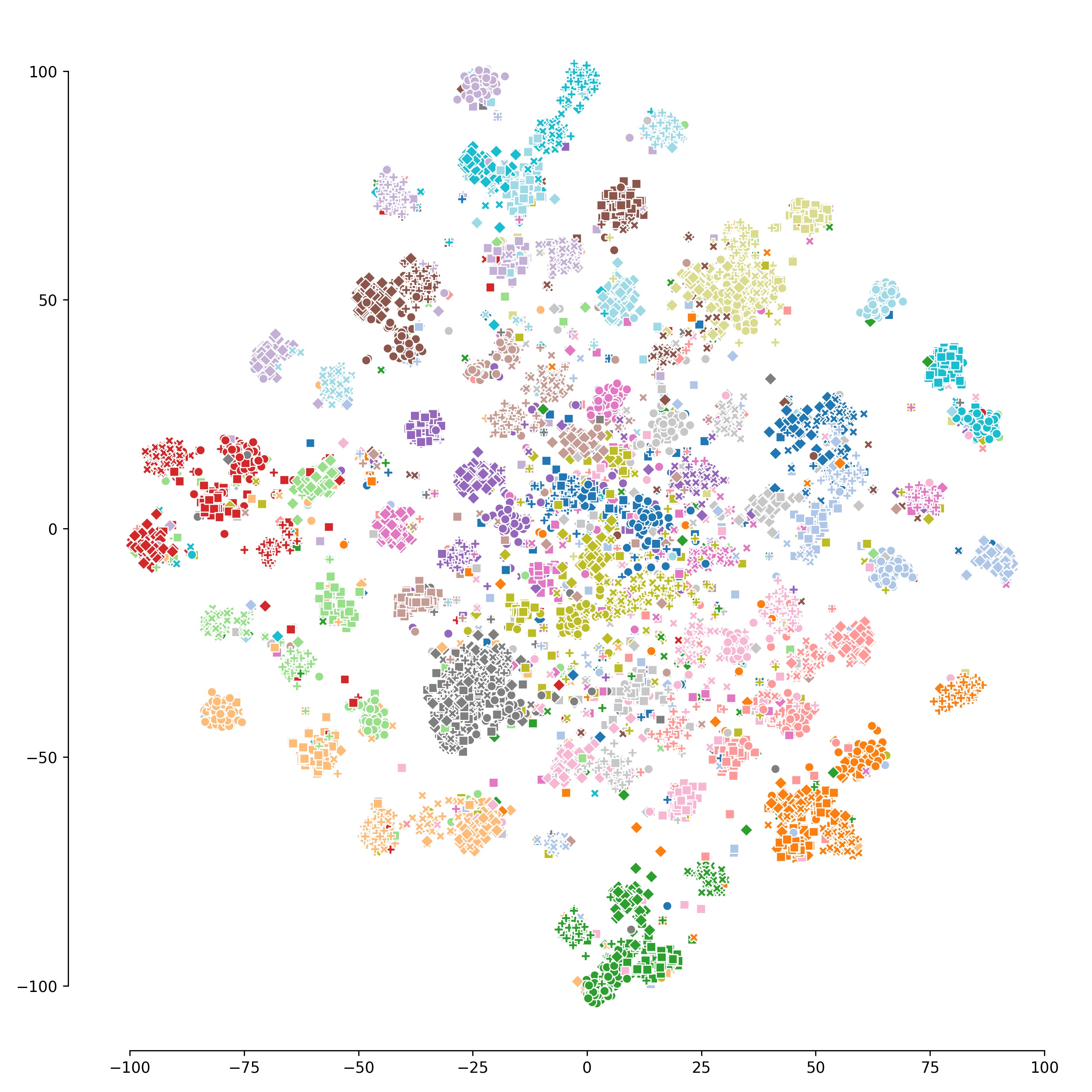}
  \caption{\RINCEout}
  \label{fig:tsne2_crossentropy}
\end{subfigure} 
  \hfill
\begin{subfigure}{0.49\textwidth}
  \centering
  \includegraphics[width=1\linewidth]{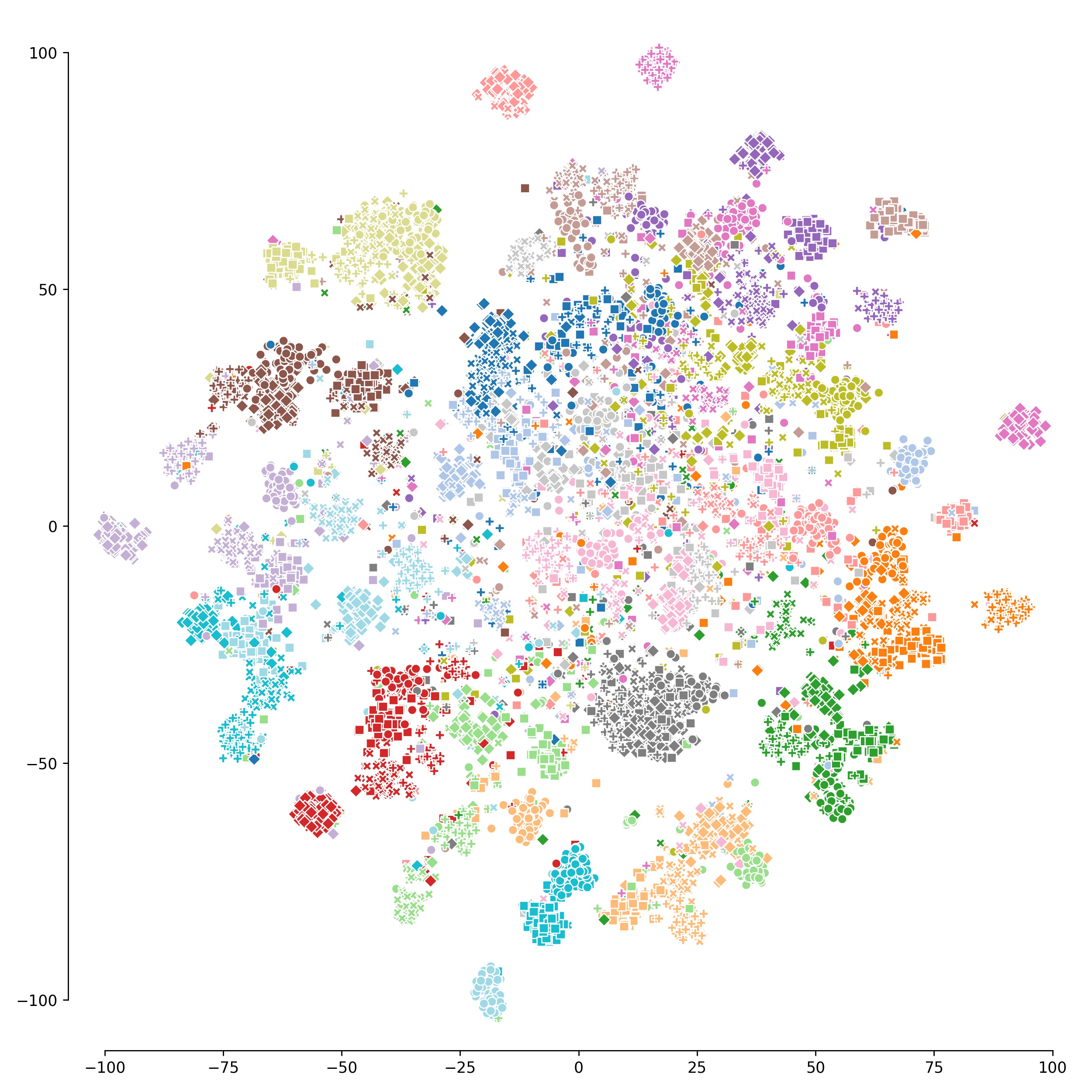}
  \caption{Cross-entropy strong aug.}
  \label{fig:tsne2_sclsuperclass}
\end{subfigure} 

\caption{Alternative version of Fig.~\ref{fig:tsne}. Here superclasses have the same color and within a superclass the marker type denotes the class label (best viewed on screen and zoomed in). T-sne plot \cite{van2008visualizing} of entire CIFAR-100 test set. For (a) supervised contrastive learning and (b) \RINCEoutin. (c) \RINCEout. (d) Cross-entropy strong aug.}
\label{fig:tsne2}
\end{figure*}

\begin{figure*}[t]
\centering
    \begin{subfigure}{0.48\textwidth}
        \centering
        \includegraphics[width=1\linewidth]{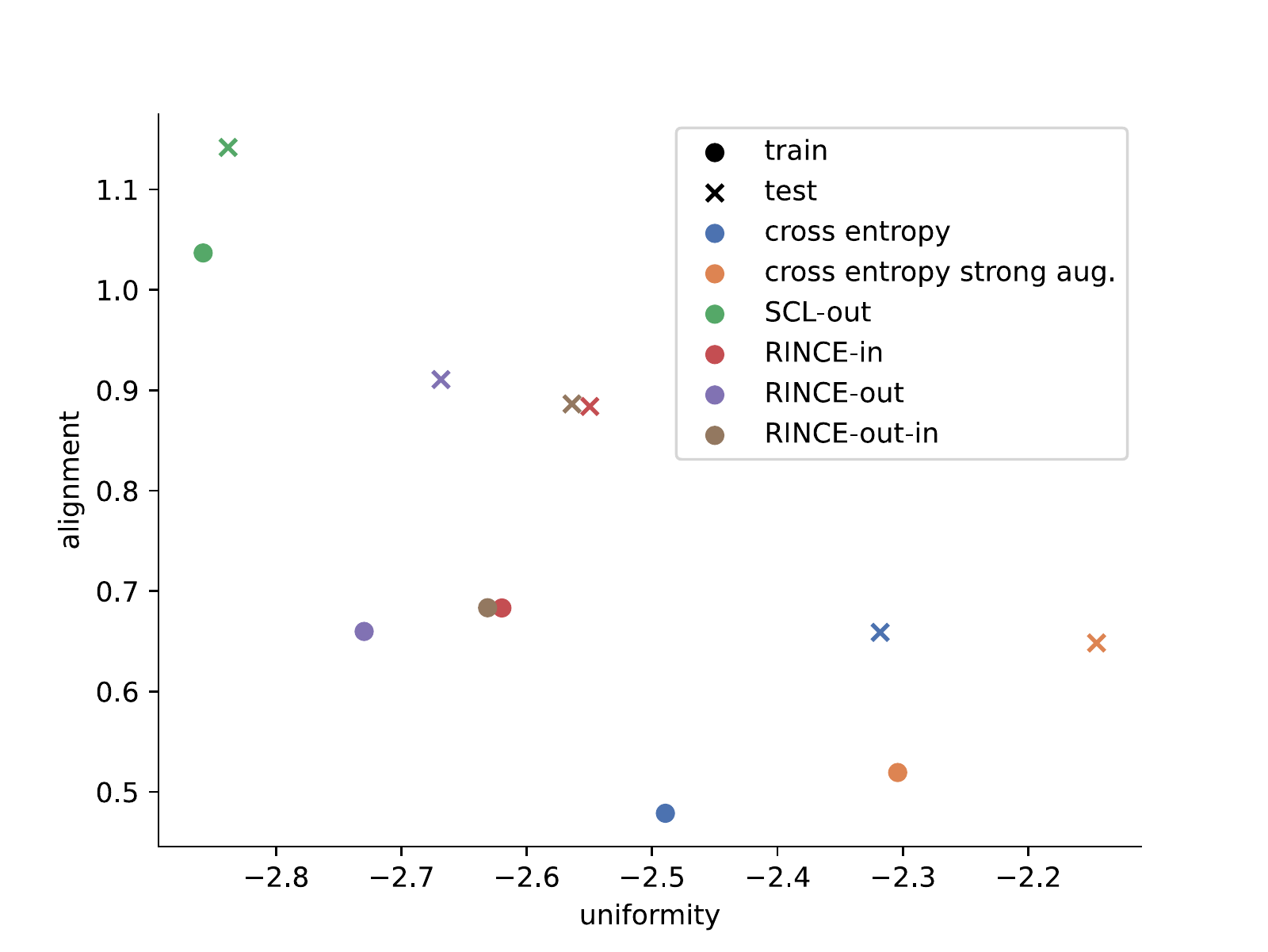}
        \caption{}
        \label{fig:unif_align_plot}
          \hfill
    \end{subfigure}        
\centering
    \begin{subfigure}{0.48\textwidth}
        \centering
        \includegraphics[width=1\linewidth]{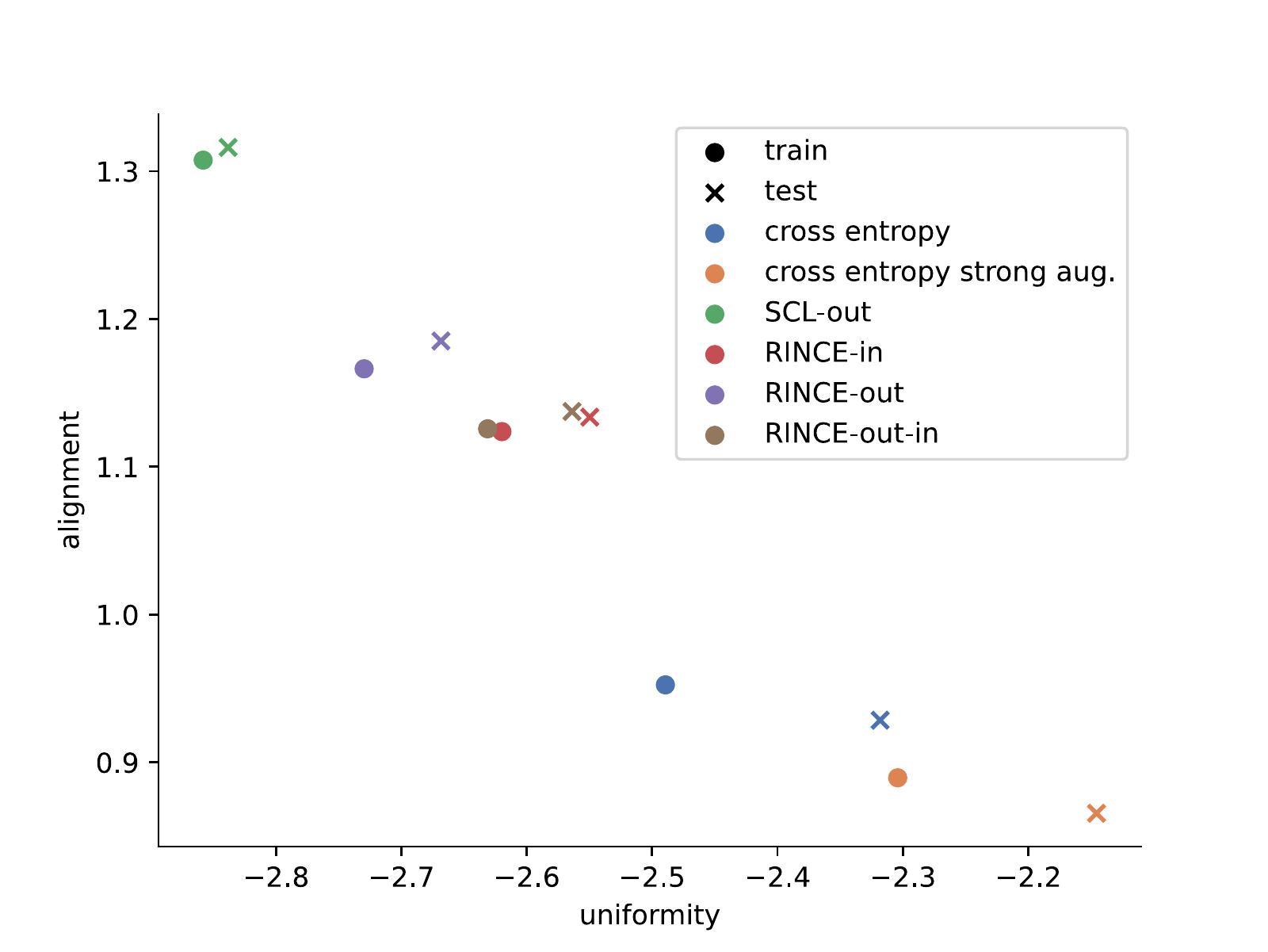}
        \caption{}
        \label{fig:unif_align_plot_coarse}   
    \end{subfigure}
    \caption{Alignment and uniformity for various models trained on \cifarhun. Methods with both, low alignment and low uniformity scores tend to generalize better to new downstream tasks \cite{wang2020understanding}, thus best models are close to the bottom left. Alignment and uniformity computed on \cifarhun train and test data for (a) fine labels and (b) for coarse labels. We compute alignment and uniformity before the MLP.}
    \label{fig:unif_align_full}
\end{figure*}

The experiments in the main paper show that RINCE leads to emergence of a well structured embedding space with useful properties, but whether RINCE learns to rank was evaluated only indirectly. To empirically validate that our method preserves the desired ranking we compute the average cosine distance between classes. If ranking is encoded in the embedding, highest similarity should be seen for same class, second highest for samples from the superclasses and low similarity for others. We visualize the results as similarity matrix in Fig.~\ref{fig:confusion_mat}. It can be seen that \RINCEin does learn the ranking to some extent in the embedding space (Fig.~\ref{fig:conf_mat_rince_in_before}). Differences become more apparent after the MLP head. Here superclasses show high similarity, which however, does not approach the within-class-similarity (compare Fig.~\ref{fig:conf_mat_rince_in_before} and \ref{fig:conf_mat_rince_in_after}). This shows that ranking is implemented to a large extent in the MLP. SCL-in also tends to have slightly higher similarity within superclasses (Fig.\ref{fig:conf_mat_scl_in_before}), but after the MLP mostly within-class-similarity is preserved (Fig.\ref{fig:conf_mat_scl_in_after}). This observation confirms that Eq.~\eqref{eq:rince_in} mirrors the desired ranking in the latent space. Enforcing such a structure in the output space, results in a better representation in the intermediate layer, as the previous layer should explore the underlying structure of data more intensively to provide the last layers of low capacity with enough information for the ranking.

\paragraph{T-SNE.}

Fig.~\ref{fig:tsne2}, depicts t-SNE plots. Color of the points denotes the superclasses of \cifarhun. Similar to the findings discussed in the previous section, it can be seen that SCL (Fig.~\ref{fig:tsne_SCL}) does a relatively bad job clustering the superclasses together, while RINCE and cross-entropy tend to group same superclasses together. 

\subsection{Alignment and Uniformity}
Another way to study the representations learned with RINCE is by examining how it influences the common alignment and uniformity metrics \cite{wang2020understanding}. As shown in \cite{wang2020understanding}, low alignment scores in combination with low uniformity scores correlate with high downstream task performance. We find that training with RINCE results in a better trade-off between alignment and uniformity, compared to cross-entropy and SCL (see Fig.~\ref{fig:unif_align_plot}).

Training with RINCE extends the alignment property across multiple samples. Therefore, positives are close to other rank 1 positives (standard alignment) but positives are also relatively close to rank 2 positives without sacrificing too much uniformity (compare Fig.~\ref{fig:unif_align_plot_coarse}). For normal \infoNCE the rank 2 positives can be very far, as can be seen by poor alignment for SCL in Fig.~\ref{fig:unif_align_plot_coarse}.

\paragraph{Connection to OOD.}
Intuitively, very low uniformity (close to uniform distribution) will result in very low OOD detection. Too high alignment, on the other hand can only be achieved, by ignoring many features. These features might be important to either spot OOD samples or generalize to similar, unseen samples. As a result, to achieve high OOD accuracy with a density estimation based approach a good trade-off between alignment and uniformity must be found.

\end{document}